\documentclass{article}

\PassOptionsToPackage{numbers, compress}{natbib}
\usepackage[final]{neurips_2021}

\usepackage[utf8]{inputenc} 
\usepackage[T1]{fontenc}    
\usepackage{hyperref}       
\usepackage{url}            
\usepackage{booktabs}       
\usepackage{amsfonts}       
\usepackage{nicefrac}       
\usepackage{microtype}      
\usepackage{xcolor}         
\usepackage{graphicx}
\usepackage{amsmath}
\usepackage{amssymb}
\usepackage{listings}
\usepackage{multirow}

\DeclareMathOperator*{\argmin}{arg\,min}

\newcommand{\algName}{EvoGrad}

\usepackage{algorithm}
\usepackage{algorithmic}

\definecolor{codegreen}{rgb}{0,0.6,0}
\definecolor{codegray}{rgb}{0.5,0.5,0.5}
\definecolor{codepurple}{rgb}{0.58,0,0.82}
\definecolor{backcolour}{rgb}{0.97,0.97,0.97}

\lstdefinestyle{mystyle}{
    backgroundcolor=\color{backcolour},   
    commentstyle=\color{codegreen},
    keywordstyle=\color{codegreen},
    numberstyle=\tiny\color{codegray},
    stringstyle=\color{codepurple},
    basicstyle=\fontsize{8}{9}\selectfont\ttfamily,
    breakatwhitespace=false,         
    breaklines=true,                 
    captionpos=b,                    
    keepspaces=false,          
    showspaces=false,                
    showstringspaces=false,
    showtabs=false,                  
    tabsize=2
}

\title{EvoGrad: Efficient Gradient-Based Meta-Learning and Hyperparameter Optimization}

\author{%
  Ondrej Bohdal$^1$, Yongxin Yang$^1$, Timothy Hospedales$^{1,2}$ \\
  $^1$ School of Informatics, The University of Edinburgh \\
  $^2$ Samsung AI Research Centre, Cambridge\\
  \texttt{\{ondrej.bohdal, yongxin.yang, t.hospedales\}@ed.ac.uk} \\
}

\begin{document}

\maketitle

\begin{abstract}
 Gradient-based meta-learning and hyperparameter optimization have seen significant progress recently, enabling practical end-to-end training of neural networks together with many hyperparameters. Nevertheless, existing approaches are relatively expensive as they need to compute second-order derivatives and store a longer computational graph. This cost prevents scaling them to larger network architectures. We present EvoGrad, a new approach to meta-learning that draws upon evolutionary techniques to more efficiently compute hypergradients. EvoGrad estimates hypergradient with respect to hyperparameters without calculating second-order gradients, or storing a longer computational graph, leading to significant improvements in  efficiency. We evaluate EvoGrad on three substantial recent meta-learning applications, namely cross-domain few-shot learning with feature-wise transformations, noisy label learning with Meta-Weight-Net and low-resource cross-lingual learning with meta representation transformation. The results show that EvoGrad significantly improves efficiency and enables scaling meta-learning to bigger architectures such as from ResNet10 to ResNet34.
\end{abstract}

\section{Introduction}

Gradient-based meta-learning and hyperparameter optimization have been of long-standing interest in neural networks and machine learning \citep{Larsen1996DesignSet, Maclaurin2015Gradient-basedLearning, Bengio2000Gradient-basedHyperparameters}. Hyperparameters (aka meta-parameters) can take diverse forms, especially under the guise of meta-learning, where there has recently been an explosion of successful applications addressing diverse learning challenges \cite{Hospedales2021Meta-learningSurvey}. For example to name just a few: training optimizer initial condition in support of few-shot learning \cite{Finn2017Model-agnosticNetworks, Antoniou2019HowMAML, Li2017Meta-SGD:Learning}; training instance-wise weights for cleaning noisy datasets \cite{Shu2019Meta-Weight-Net:Weighting, Ren2018LearningLearning}; training loss functions in support of generalisation  \cite{Li2019Feature-criticGeneralization} and learning speed; and training stochastic regularizers in support of cross-domain robustness \cite{Tseng2020Cross-domainTransformation}.

Most of these applications share the property that meta-parameters impact validation loss only indirectly through their effect on model parameters, and so computing validation loss gradients with respect to meta-parameters usually leads to the need to compute second-order derivatives, and store longer computational graphs for backpropagation. This eventually becomes a bottleneck to execution time, and -- more severely -- to scaling the size of the underlying models, given the practical limitation of GPU memory. There has been steady progress in the development of diverse practical algorithms for computing validation loss with respect to meta-parameters \cite{Luketina2016ScalableHyperparameters,Lorraine2020OptimizingDifferentiation,Maclaurin2015Gradient-basedLearning}. Nevertheless they mostly share some form of the aforementioned limitations. In particular, the  majority of recent successful practical applications \cite{Shu2019Meta-Weight-Net:Weighting,Tseng2020Cross-domainTransformation,Li2019Feature-criticGeneralization, Balaji2018MetaReg:Meta-regularization, Bohdal2020FlexibleImages, Liu2019Self-SupervisedLearning, Shan2020Meta-Neighborhoods} essentially use some variant of the $T_1-T_2$ algorithm \cite{Luketina2016ScalableHyperparameters} to estimate the gradient $\frac{\partial\ell_V}{\partial\lambda}$ of validation loss w.r.t. hyperparameters. This approach computes the gradient online at each step of updating the base model $\theta$, and estimates it as $\frac{\partial\ell_V}{\partial\lambda}\approx\frac{\partial\ell_V}{\partial\theta}\frac{\partial^2\ell_T}{\partial\theta\partial\lambda}$, for training loss $\ell_T$. As with many alternative estimators, this requires second-order derivatives, and extending the computational graph. Besides the additional computation cost, this limits the size of the base model that can be used in a given GPU, since the memory cost of meta-learning is now multiple times the size of vanilla backpropagation. This in turn prevents the application of meta-learning to problems where large state-of-the-art model architectures are required. 

To address this issue, we draw inspiration from evolutionary optimization methods \cite{salimans2017evolution} to develop \algName{}, a meta-gradient algorithm that requires no higher-order derivatives and as such is significantly faster and lighter than the standard approaches. In particular, we take the novel view of estimating meta-gradients via a putative inner-loop \emph{evolutionary} update to the base model. As this requires no gradients itself, the meta-gradient can then be computed using first-order gradients alone, and without extending the computational graph -- leading to efficient hyperparameter updates. Meanwhile for efficient and accurate base model learning, the real inner-loop update can separately be carried out by conventional gradient descent.

Our \algName{} is a general meta-optimizer applicable to many meta-learning applications, among which we choose three to demonstrate its impact: the LFT model \cite{Tseng2020Cross-domainTransformation} observes that a properly tuned stochastic regularizer can significantly improve cross-domain few-shot learning performance. We show that by training those regularizer parameters with \algName{}, rather than the standard second-order approach, we can obtain the same improvement in accuracy with significant reduction in time and memory cost. This allows us to scale LFT from the original ResNet10 to ResNet34 within a 12GB GPU. Second, the Meta-Weight-Net (MWN) \cite{Shu2019Meta-Weight-Net:Weighting} model deals with label noise by meta-learning an auxiliary network that re-weights instance-wise losses to down-weight noisy instances and improve validation loss. We also show that EvoGrad can replicate MWN results with significant cost savings. Third, we demonstrate the benefits of EvoGrad on an application from NLP, in addition to the ones from computer vision: low-resource cross-lingual learning using MetaXL approach \citep{Xia2021MetaXL:Learning}.

To summarize, our main contributions are: (1) We introduce \algName{}, a novel method for gradient-based meta-learning and hyperparameter optimization that is simple to implement and efficient in time and memory requirements. (2) We evaluate \algName{} on a variety of illustrative and substantial meta-learning problems, where we demonstrate significant compute and memory benefits compared to standard second-order approaches. (3) In particular, we illustrate that \algName{} allows us to scale meta-learning to bigger models than was previously possible on a given GPU size, thus bringing meta-learning closer to the state-of-the-art frontier of real applications. We provide source code for EvoGrad at \url{https://github.com/ondrejbohdal/evograd}.

\section{Related work}
Gradient-based meta-learning solves a bilevel optimization problem where validation loss is optimized with respect to the meta-knowledge by backpropagating through the update of the model on training data and with meta-knowledge. The meta-knowledge updates form an outer loop, around an inner loop of base model updates. The inner loop can run for one \cite{Luketina2016ScalableHyperparameters}, few \cite{Shaban2019TruncatedOptimization,Maclaurin2015Gradient-basedLearning}, or many \cite{Lorraine2020OptimizingDifferentiation} steps within each outer-loop iteration. 
Meta-knowledge can take many forms, for example, it can be an initialization of the model weights \citep{Finn2017Model-agnosticNetworks}, feature-wise transformation layers \citep{Tseng2020Cross-domainTransformation}, regularization to improve domain generalization \citep{Balaji2018MetaReg:Meta-regularization} or even a synthetic training set \citep{Wang2018DatasetDistillation, Bohdal2020FlexibleImages}. 
Most substantial practical applications use a one or few-step inner loop for efficiency.

More recently, several methods \citep{Lorraine2020OptimizingDifferentiation, Rajeswaran2019Meta-learningGradients} have utilized Implicit Function Theorem (IFT) to develop new gradient-based meta-learners. These methods use multiple inner-loop steps without the need to backpropagate through whole inner loop, which significantly improves memory efficiency over methods that need keep track of the whole inner loop training process. However, IFT methods assume the model has converged in the inner loop. This makes them unsuited for the majority of practical applications above where training the inner loop to convergence for each hypergradient step is infeasible. Furthermore, the hypergradient is still more costly compared to one-step $T_1-T_2$  method. The costs come from the associated overhead with approximating an inverse Hessian of the training data with respect to the model parameters. Note that the Hessian itself does not need to be stored due to the mechanics of reverse-mode differentiation \citep{Griewank1993SomeHessians, Baydin2018AutomaticSurvey}. However, this does not eliminate the remaining calculations which still require higher-order gradients that result in backpropagation via longer graphs due to additional gradient nodes. For these reasons, we focus comparison on the more widely used $T_1-T_2$ strategy which is oriented at single-step inner loops similar to \algName{}.

Theoretically it is also possible to use hypernetworks \citep{Lorraine2018StochasticHypernetworks} to find good hyperparameters in a first-order way. Hypernetworks take hyperparameters as inputs and generate model parameters. However, the approach is not commonly used, likely due to the difficulty of generating well-performing model parameters. We provide experimental results to support this hypothesis in the supplementary material.

Meta-learning can be categorized into several groups, depending on the type of meta-knowledge and also if the model is trained from scratch as part of the inner loop \citep{Hospedales2021Meta-learningSurvey}. Offline meta-learning approaches train a model from scratch per each update of the meta-knowledge, while online meta-learning approaches train the model and meta-knowledge jointly. As a result, offline meta-learning is extremely expensive \cite{cubuk2019autoAug,zoph2017neuralArchitectureRL} when scaled beyond few-shot learning problems where only a few iterations are sufficient for training \citep{Finn2017Model-agnosticNetworks, Antoniou2019HowMAML, Li2017Meta-SGD:Learning}. Therefore most larger-scale problems \cite{liu2019darts,jaderberg2019popRL,Tseng2020Cross-domainTransformation, Xia2021MetaXL:Learning} use online learning in practice, and this is where we focus our contribution.

The meta-knowledge to learn can take different forms. A particular dichotomy is between the special case where the meta-knowledge corresponds to the base model itself, in the form of an initialization; and the more general cases where it does not. The former initialization meta-learning has been popularized by MAML \citep{Finn2017Model-agnosticNetworks}, and is widely used in few-shot learning. This can be solved relatively efficiently, for example using a first-order approximation of MAML \citep{Finn2017Model-agnosticNetworks}, Reptile \citep{Nichol2018OnAlgorithms} or minibatch proximal update \citep{Zhou2019EfficientUpdate}. On the other hand, there are vastly more cases \cite{Hospedales2021Meta-learningSurvey} where the meta-knowledge is different from the model itself, such as LFT's stochastic regularizer to improve cross-domain generalization \citep{Tseng2020Cross-domainTransformation}, MWN's  instance-wise loss weighting network for label noise robustness \citep{Shu2019Meta-Weight-Net:Weighting}, a label generation network to improve self-supervised generalization \citep{Liu2019Self-SupervisedLearning}, a Feature-Critic loss  to improve domain generalization \citep{Li2019Feature-criticGeneralization} and many others.
In this more general case, most applications rely on a $T_1-T_2$-like algorithm, as the efficient approximations specific to MAML do not apply. 
The ability to significantly improve the efficiency of gradient-based meta-learning would have a large impact as methods like these would directly benefit from it in runtime and energy consumption. More crucially, they could scale to bigger and more state-of-the-art neural network architectures.

\begin{figure}[t]
  \centering
  \includegraphics[width=\textwidth]{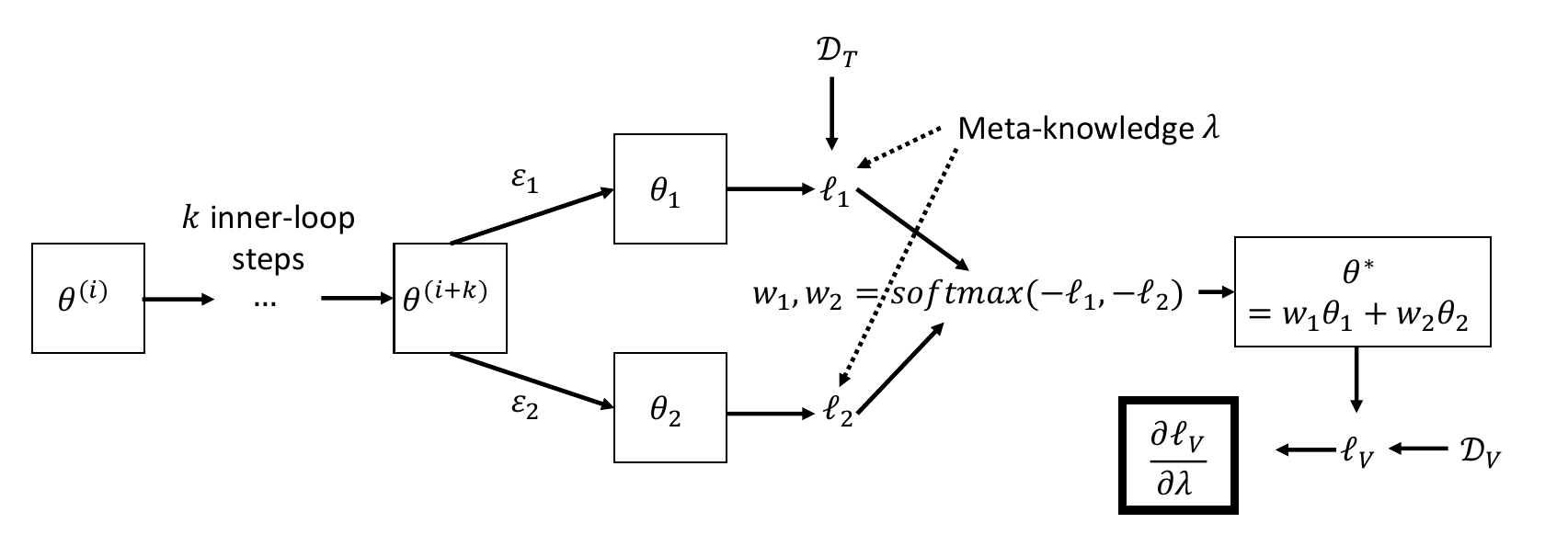}
  \caption{Graphical illustration of a single \algName{} update using $K=2$ model copies.}
  \label{fig:evoGradDiagram}
\end{figure}

\section{Methods}
\subsection{Background: meta-learning as bilevel optimization}
We aim to solve a bilevel optimization problem where our goal is to find hyperparameters $\boldsymbol{\lambda}$ that minimize the validation loss $\ell_V$ of the model parametrized by $\boldsymbol{\theta}$ and trained with loss $\ell_T$ and $\boldsymbol{\lambda}$:

\begin{equation}
\boldsymbol{\lambda}^*=\argmin_{\boldsymbol{\lambda}}\ell_V^*( \boldsymbol{\lambda}), \text{ where } \ell_V^*(\boldsymbol{\lambda}) = \ell_V(\boldsymbol{\lambda}, \boldsymbol{\theta}^*(\boldsymbol{\lambda})) \text{ and } \boldsymbol{\theta}^*(\boldsymbol{\lambda)} = \argmin_{\boldsymbol{\theta}} \ell_T(\boldsymbol{\lambda}, \boldsymbol{\theta}).
\end{equation}

In order to meta-learn the value of $\boldsymbol{\lambda}$ using gradient-based methods, we need to calculate the hypergradient $\frac{\partial \ell_V}{\partial \boldsymbol{\lambda}}$. We can expand its calculation as follows:
\begin{equation}
\frac{\partial \ell_V^*(\boldsymbol{\lambda})}{\partial \boldsymbol{\lambda}} = \frac{\partial \ell_V(\boldsymbol{\lambda}, \boldsymbol{\theta}^*(\boldsymbol{\lambda}))}{\partial \boldsymbol{\lambda}} + \frac{\partial \ell_V(\boldsymbol{\lambda}, \boldsymbol{\theta}^*(\boldsymbol{\lambda}))}{\partial \boldsymbol{\theta}^*(\boldsymbol{\lambda})}\frac{\partial \boldsymbol{\theta}^*(\boldsymbol{\lambda})}{\partial \boldsymbol{\lambda}}.
\end{equation}

In meta-learning and hyperparameter optimization more broadly, the direct term $\frac{\partial \ell_V(\boldsymbol{\lambda}, \boldsymbol{\theta}^*(\boldsymbol{\lambda}))}{\partial \boldsymbol{\lambda}}$ is typically zero because the hyperparameter does not directly influence the value of the validation loss -- it influences it via the impact on the model weights $\boldsymbol{\theta}$. However, the model weights $\boldsymbol{\theta}$ are themselves trained using gradient optimization, which gives rise to higher-order derivatives. We propose a variation on this step where the update of the model weights is inspired by evolutionary methods, allowing us to eliminate the need for higher-order derivatives. We consider the setting where the hypergradient of hyperparameter $\lambda$ is estimated online \cite{Luketina2016ScalableHyperparameters} together with updating the base model $\theta$, as this is the most widely used setting in substantial practical applications \cite{Shu2019Meta-Weight-Net:Weighting,Tseng2020Cross-domainTransformation,Li2019Feature-criticGeneralization, Balaji2018MetaReg:Meta-regularization, Bohdal2020FlexibleImages, Shan2020Meta-Neighborhoods,liu2019darts, Xia2021MetaXL:Learning}.

\subsection{The EvoGrad update}
Given the current model parameters $\theta \in \mathbb{R}^M$, hyperparameters  $\lambda \in \mathbb{R}^N$, training loss $\ell_{T}$ and validation loss $\ell_{V}$, we aim to estimate $\frac{\partial \ell_{V}}{\partial \boldsymbol{\lambda}}$ for efficient gradient-based hyperparameter learning. The key idea is -- solely for the purpose of hypergradient estimation -- to consider a simple evolutionary rather than gradient-based inner-loop step on $\theta$.

\textbf{Evolutionary inner step}\quad 
First, we sample random perturbations $\boldsymbol{\epsilon} \in \mathbb{R}^M \sim \mathcal{N}(\boldsymbol{0}, \sigma \boldsymbol{I})$, and apply them to $\boldsymbol{\theta}$. Sampling $K$ perturbations, we can create a population of $K$ variants $\{\boldsymbol{\theta}_k\}_{k=1}^K$ of the current model as  $\boldsymbol{\theta}_k = \boldsymbol{\theta}+\boldsymbol{\epsilon}_k$. We can now compute the training losses $\{\ell_k\}^K_{k=1}$ for each of the $K$ models, $\ell_k=f(\mathcal{D}_T|\boldsymbol{\theta}_k,\boldsymbol{\lambda})$ using the current minibatch $\mathcal{D}_T$ drawn from the training set. Given these loss values, we can calculate the weights (sometimes called fitness) of the population of candidate models as

\begin{equation}
w_1, w_2, \dots, w_K = \operatorname{softmax}([-\ell_1, -\ell_2, \dots, -\ell_K]/\tau),
\label{eq:weight}
\end{equation}
where $\tau$ is a temperature parameter that rescales the losses to control the scale of weight variability.

Given the weights $\{w_k\}_{k=1}^K$, we complete the current step of evolutionary learning by updating the model parameters via the affine combination
\begin{equation}
\boldsymbol{\theta}^* = w_1\boldsymbol{\theta}_1 + w_2\boldsymbol{\theta}_2 + \dots + w_K\boldsymbol{\theta}_K.
\end{equation}

\textbf{Computing the hypergradient}\quad We now evaluate the updated model $\boldsymbol{\theta}^*$ for a minibatch from the validation set $\mathcal{D}_V$  
and take gradient of the validation loss $\ell_{V} = f(\mathcal{D}_V | \boldsymbol{\theta}^*)$ w.r.t. the hyperparameter:
\begin{equation}
\frac{\partial \ell_{V}}{\partial \boldsymbol{\lambda}}=
\frac{\partial f(\mathcal{D}_V | \boldsymbol{\theta}^*)}{\partial \boldsymbol{\lambda}}
\label{eq:grad}
\end{equation}
One can easily verify that the computation in Eq.~\ref{eq:grad} does not involve second-order gradients as no first-order gradients were used in the inner loop. This is in contrast to the typical approach \cite{Luketina2016ScalableHyperparameters,Maclaurin2015Gradient-basedLearning} of applying gradient-based updates in the inner loop and differentiating through it (in either forward-mode or reverse-mode), or even applying the implicit function theorem (IFT) \cite{Lorraine2020OptimizingDifferentiation}, all of which trigger higher-order gradients and an extended computation graph.

\textbf{Algorithm flow}\quad In practice we follow the flow of $T_1-T_2$ \cite{Luketina2016ScalableHyperparameters} used by many substantive applications \cite{Tseng2020Cross-domainTransformation,Shu2019Meta-Weight-Net:Weighting,Balaji2018MetaReg:Meta-regularization,liu2019darts, Xia2021MetaXL:Learning}. We take alternating steps on $\theta$ using the exact gradient $\frac{\partial\ell_{T}}{\partial\boldsymbol\theta}$, and on $\boldsymbol\lambda$ using the hypergradient 
$\frac{\partial \ell_{V}}{\partial \boldsymbol{\lambda}}$, which in \algName{} is estimated  as in Eq.~\ref{eq:grad}. 

\subsection{EvoGrad hypergradient as a random projection}\label{sec:project}
To understand \algName{}, observe that the hyper-gradient in Eq.~\ref{eq:grad} expands as
\begin{equation}
\frac{\partial \ell_{V}}{\partial \boldsymbol{\lambda}} = \frac{\partial \ell_{V}}{\partial \boldsymbol{\theta}^*} \frac{\partial \boldsymbol{\theta}^*}{\partial \boldsymbol{\lambda}} = \frac{\partial \ell_{V}}{\partial \boldsymbol{\theta}^*} \mathcal{E} \frac{\partial \boldsymbol{w}}{\partial \boldsymbol{\lambda}}
= \frac{\partial \ell_{V}}{\partial \boldsymbol{\theta}^*} \mathcal{E}  \frac{\partial \boldsymbol{w}}{\partial \boldsymbol{\ell}} \frac{\partial  \boldsymbol{\ell}}{\partial \boldsymbol{\lambda}}
\end{equation}

where $\mathcal{E} = [\boldsymbol{\epsilon}_1, \boldsymbol{\epsilon}_2, \dots, \boldsymbol{\epsilon}_K]$ is the  $M\times K$ matrix formed by stacking $\boldsymbol{\epsilon}_k$'s as columns, $\boldsymbol{w}$ is the $K$-dimensional ($\boldsymbol w=[w_1,w_2,\dots,w_K]$) vector of candidate model weights, and $\boldsymbol{\ell} = [\ell_1,\ell_2,\dots,\ell_K]$  is the $K$-dimensional vector of candidate model losses. 

Recall that $\mathcal{E}$ is a random matrix, so the operation $\frac{\partial \ell_{V}}{\partial \boldsymbol{\theta}^*} \mathcal{E}$ can be understood as randomly projecting the $M$-dimensional validation loss' gradient to a new low-dimensional space of dimension $K\ll M$. 
Alternatively, we can interpret the update as factorising the model-parameter-to-hyperparameter derivative $\frac{\partial \boldsymbol{\theta}^*}{\partial \boldsymbol{\lambda}}$ (sized $M\times N$) into two much smaller matrices $\mathcal{E}$ and $\frac{\partial \boldsymbol{w}}{\partial \boldsymbol{\lambda}}$ of size $M\times K$ and $K\times N$.

In terms of implementation, $\frac{\partial \ell_{V}}{\partial \boldsymbol{\theta}^*}$ is obtained by backpropagation and $\mathcal{E}$ is sampled on the fly. The term $\frac{\partial w}{\partial \boldsymbol{\lambda}} = \frac{\partial \boldsymbol{w}}{\partial \boldsymbol{\ell}} \frac{\partial  \boldsymbol{\ell}}{\partial \boldsymbol{\lambda}}$ is computed by the softmax-to-logit derivative ($K\times K)$ and the derivative of the $K$ candidate models training losses w.r.t. hyperparameters. It is noteworthy that the $K$ elements of $\frac{\partial  \boldsymbol{\ell}}{\partial \boldsymbol{\lambda}}$ are completely independent, and can be computed in parallel where multiple GPUs are available.

\subsection{Comparison to other methods}
We compare \algName{} to the most related and widely-used alternative $T_1-T_2$ \citep{Luketina2016ScalableHyperparameters} in Table~\ref{tab:hypergradcomparison}. $T_1-T_2$ requires higher-order gradients and associated longer computational graphs -- due to the need to backpropagate through gradient nodes. This leads to increased memory and time cost compared to vanilla backpropagation. In contrast, \algName{} requires no higher-order gradients, no large matrices, and no substantial expansion of the computational graph. 

\begin{table}[h!]
\caption{Comparison of hypergradient approximations of $T_1-T_2$ and EvoGrad. }
\label{tab:hypergradcomparison}
\centering
\begin{tabular}{ll}
\toprule
Method & Hypergradient approximation \\
\midrule
$T_1-T_2$ \citep{Luketina2016ScalableHyperparameters}&  $\frac{\partial \ell_V}{\partial \boldsymbol{\lambda}}-\frac{\partial \ell_V}{\partial \boldsymbol{\theta}} \times \boldsymbol{I} \frac{\partial^{2} \ell_T}{\partial \boldsymbol{\theta} \partial \boldsymbol{\lambda}^{T}}$\\ 
EvoGrad (ours) & $\frac{\partial \ell_V}{\partial \boldsymbol{\lambda}}+\frac{\partial \ell_{V}}{\partial \boldsymbol{\theta}} \times \mathcal{E}  \frac{\partial \boldsymbol{w}}{\partial \boldsymbol{\ell}} \frac{\partial  \boldsymbol{\ell}}{\partial \boldsymbol{\lambda}} = \frac{\partial \ell_V}{\partial \boldsymbol{\lambda}}+\frac{\partial \ell_V}{\partial \boldsymbol{\theta}} \times \mathcal{E} \frac{\partial \operatorname{softmax}(-\boldsymbol{\ell})}{\boldsymbol{\partial\lambda}}$ \\
\bottomrule
\end{tabular}
\end{table}

We analyse the asymptotic big-$\mathcal{O}$ time and memory requirements of \algName{} vs $T_1-T_2$ in Table \ref{tab:memorycomparisonbigo}. The dominant cost in terms of both memory and time is the cost of backpropagation. Backpropagation is significantly more expensive than forward propagation because forward propagation does not need to store all intermediate variables in memory \citep{Rajeswaran2019Meta-learningGradients, Griewank1993SomeHessians}. Note that even if \algName{} keeps multiple copies of the model weights in memory, this cost is small compared to the cost of backpropagation, and the latter is done with only \emph{one} set of weights $\boldsymbol\theta^*$. We remark that our main empirical results are obtained with only $K=2$ models, so we can safely ignore this in our asymptotic analysis.

In addition, we elaborate on how higher-order gradients contribute to increased memory and time costs. Results from computing the first-order gradients are added into the computation graph as new nodes in the graph so that we can calculate the higher-order gradients. When calculating the higher-order gradients, we backpropagate through this longer computational graph, which directly increases the memory and time costs. The current  techniques \cite{Luketina2016ScalableHyperparameters} rely on longer computational graphs, while \algName{} significantly shortens the graph and reduces memory cost by avoiding this step. This consideration is not visible in the big-$\mathcal{O}$ analysis, but contributes to improved efficiency.

\begin{table}[h!]
\caption{Comparison of asymptotic memory and operation requirements of EvoGrad and $T_1-T_2$ meta-learning strategies. $P$ is the number of model parameters, $H$ is the number of hyperparameters. 
$K\ll H$ is the number of model copies in EvoGrad. Note this is a first-principles analysis, so the time requirements are different when using e.g. reverse-mode backpropagation that uses parallelization.}
\label{tab:memorycomparisonbigo}
\centering
\begin{tabular}{lcc}
\toprule
Method & Time requirements & Memory requirements \\
\midrule
$T_1-T_2$ \citep{Luketina2016ScalableHyperparameters}& $\mathcal{O}(PH)$ & $\mathcal{O}(P+H)$ \\ 
EvoGrad (ours) & $\mathcal{O}(KP+H)$ & $\mathcal{O}(P+H)$\\
\bottomrule
\end{tabular}
\end{table}

\section{Experiments}
We first consider two simple problems: 1) a 1-dimensional problem where we try to find the minimum of a function, and 2) meta-learning a feature-transformer to find the rotation that correctly aligns images whose training and validation sets differ in rotation. This serves as a proof-of-concept problem to show our method is capable of meta-learning suitable hyperparameters. We then consider three real problems where meta-learning has been used to solve different learning challenges. We show that \algName{} makes a significant impact in terms of reducing the memory and time costs (while keeping the accuracy improvements brought by meta-learning): 3) Cross-domain few-shot classification via learned feature-wise transformation \citep{Tseng2020Cross-domainTransformation}, 4) Meta-Weight-Net: learning an explicit mapping for sample weighting \citep{Shu2019Meta-Weight-Net:Weighting}, 5) MetaXL: meta representation transformation for low-resource cross-lingual learning \citep{Xia2021MetaXL:Learning}. We provide a brief overview of each problem, together with evaluation and analysis. Further details and experimental settings are described in the supplementary material.

\subsection{Illustration using a 1-dimensional problem}
In this problem we minimize a validation loss function $f_V(x)=(x-0.5)^2$ where parameter $x$ is optimized using SGD with training loss function $f_T(x)=(x-1)^2+\lambda \|x\|_2^2$ that includes a meta-parameter $\lambda$. A closed-form solution for the hypergradient is available and is equal to $g(\lambda)=(\lambda-1)/(\lambda+1)^3$, which allows us to compare EvoGrad against the ground-truth gradient.

\begin{figure}[h!]
  \centering
  \includegraphics[width=\textwidth]{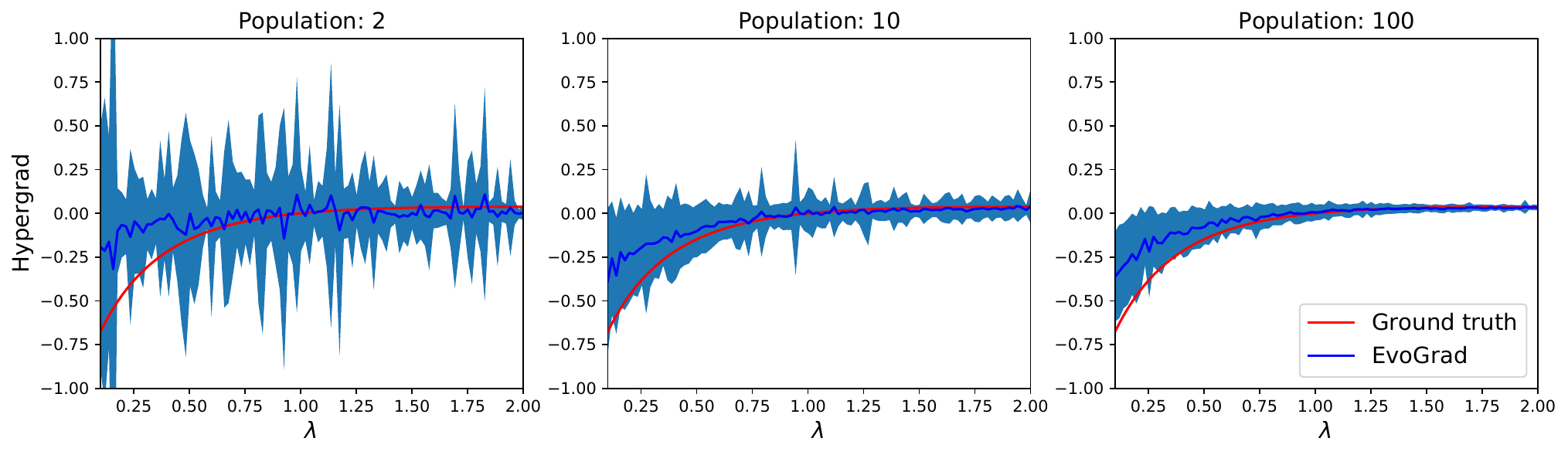}
  \vspace{-0.5cm}
  \caption{Comparison of the hypergradient $\partial f_V/\partial\lambda$ estimated by EvoGrad vs the ground-truth.}
  \label{fig:toy1dgrads}
\end{figure}

Our first analysis studies the estimated EvoGrad hypergradient for a grid of $\lambda$ values between 0 and 2. For each value of $\lambda$ we show the mean and standard deviation of the estimated $\partial f_V/\partial\lambda$ over 100 repetitions (with random choice of $x$). We use temperature $\tau=0.5$, $\epsilon \in \mathbb{R} \sim \mathcal{N}(0, 1)$ and consider between 2 and 100 models copies in the population. The results in Figure \ref{fig:toy1dgrads} show that EvoGrad estimates have a similar trend to the ground-truth gradient, even if the EvoGrad estimates are noisy. The level of noise decreases with more models in the population, but the correct trend is visible even if we use only 2 models.

\begin{figure}[h!]
  \centering
  \includegraphics[width=\textwidth]{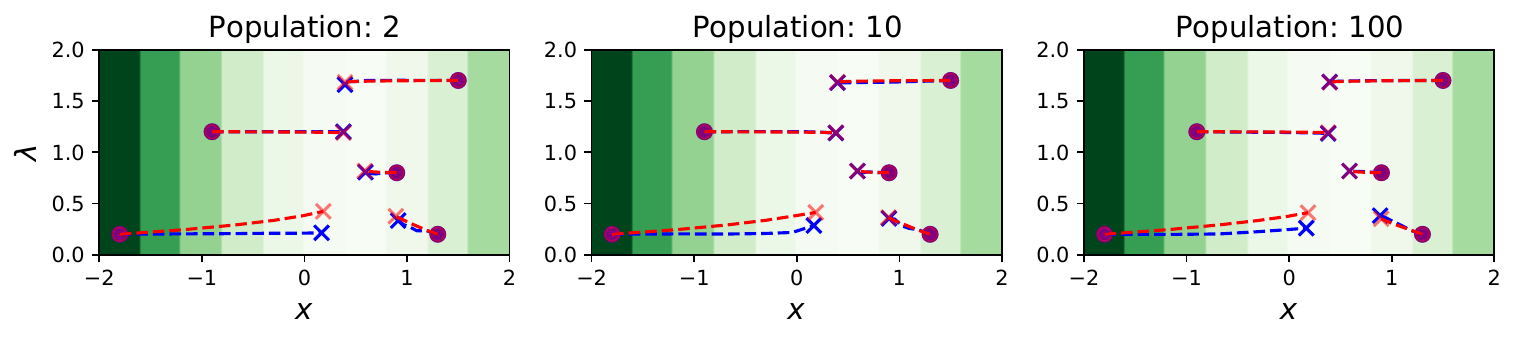}
  \vspace{-0.5cm}
  \caption{Trajectories of parameters $x, \lambda$ when following $\partial f_T/\partial x$ and $\partial f_V/\partial\lambda$ using SGD for 5 random starting positions. Comparison of trajectories using EvoGrad estimated (blue) or ground-truth (red) hypergradient. The initial position is marked with a circle, and the final position after 5 steps is marked with a cross. The shading is validation loss $f_V(x)$.}
  \label{fig:toy1dtrajs}
\end{figure}

Our second analysis studies the trajectories that parameters $x, \lambda$ follow if they are both optimized online using SGD with learning rate of 0.1 for 5 steps, starting from five different positions (circles). The hypergradients are either estimated using EvoGrad or directly using the ground-truth formula. Figure \ref{fig:toy1dtrajs} shows that the trajectories of both variations are similar, and they become more similar as we use more models in the population. In all cases the parameters converge towards the 
lightly-coloured region where the validation loss is the lowest at $x=0.5$.

\subsection{Rotation transformation}
In this task we work with MNIST images  \citep{LeCun1998MNISTDatabase}, and assume that the validation and test sets are rotated by $30^\circ$ compared to the conventionally oriented training images. Clearly, directly training a model and applying it will lead to low performance. We therefore assume meta-knowledge in the form of a hidden  rotation. The rotation transformation is applied to the training images before learning, and should itself be meta-learned by the validation loss obtained by the CNN trained on the rotated training set. Thus solving the meta-learning problem should result in a $30^\circ$ rotation, and a base CNN that generalises to the rotated validation set. 

The problem is framed as online meta-learning where each update of the base model is followed by a meta-parameter update  using EvoGrad.
We use EvoGrad with 2 model copies, temperature $\tau=0.05$ and $\sigma=0.001$ for $\boldsymbol{\epsilon} \sim \sigma \text{sign}(\mathcal{N}(\boldsymbol{0}, \boldsymbol{I}))$. Our LeNet \citep{LeCun1989HandwrittenLearning} base model is trained for 5 epochs.

We repeat the experiments 5 times and show a comparison of the results in Table \ref{tab:mnistrot}. A baseline model achieves $98.40 \pm 0.07$\% accuracy if the test images are not rotated, but its accuracy drops to $81.79 \pm 0.64$\% if the same images are rotated by $30^\circ$. A model trained with EvoGrad and the rotation transformer is able to accurately classify rotated images, with a similar accuracy as the baseline model can classify unrotated images. This confirms we can successfully optimize hyperparameters with EvoGrad. The meta-learned rotation is also close to the true value.

\begin{table}[h!]
\caption{Rotation transformation learning.  The goal is to accurately classify MNIST test images rotated by $30^\circ$ degrees compared to the  training set orientation. Test accuracies (\%) of a baseline model, and one whose training set has been rotated by the EvoGrad's meta-learned rotation, and associated EvoGrad rotation estimate ($^\circ$). Accuracy for rotation matched train/test sets is $98.40\%$.}
\label{tab:mnistrot}
\centering
\begin{tabular}{lccc}
\toprule
True Rotation & Baseline Acc. & EvoGrad Acc. & EvoGrad Rotation Est. \\
\midrule
$30^\circ$ & 81.79 $\pm$ 0.64 & 98.11 $\pm$ 0.32 & 28.47$^\circ$ $\pm$ 5.23$^\circ$ \\
\bottomrule
\end{tabular}
\end{table}

\subsection{Cross-domain few-shot classification via learned feature-wise transformation}
As the next task we consider cross-domain few-shot classification (CD-FSL). CD-FSL is considered an important and highly challenging problem at the forefront of computer vision. The state-of-the-art approach learned feature-wise transformation (LFT) \citep{Tseng2020Cross-domainTransformation} aims to meta-learn stochastic feature-wise transformation layers that regularize metric-based few-shot learners to improve their few-shot learning generalisation in cross-domain conditions. The method includes two key steps: 1) updating the model with the meta-parameters on a pseudo-seen task and 2) updating the meta-parameters by evaluating the model on a pseudo-unseen task by backpropagating through the first step. As feature-wise transformation is not directly used for the pseudo-unseen task, this leads to higher-order gradients. Note that the problem itself is memory-intensive because we work with larger images of size $224\times224$ within episodic learning tasks. As a result, a significantly more efficient meta-learning approach could allow us to scale from the ResNet10 model used in the paper to a larger model.

We experiment with the LFT-RelationNet \citep{Sung2018LearningLearning} metric-based few-shot learner and consider the exact same experiment settings as \citep{Tseng2020Cross-domainTransformation} using the official PyTorch implementation associated with the paper. LFT introduces 3712 hyper-parameters to train for ResNet10, and 9344 for ResNet34. All our experiments are conducted on Titan X GPUs with 12GB of memory using $K=2$ for \algName{}.

\begin{figure}[h!]
  \centering
  \includegraphics[width=0.9\textwidth]{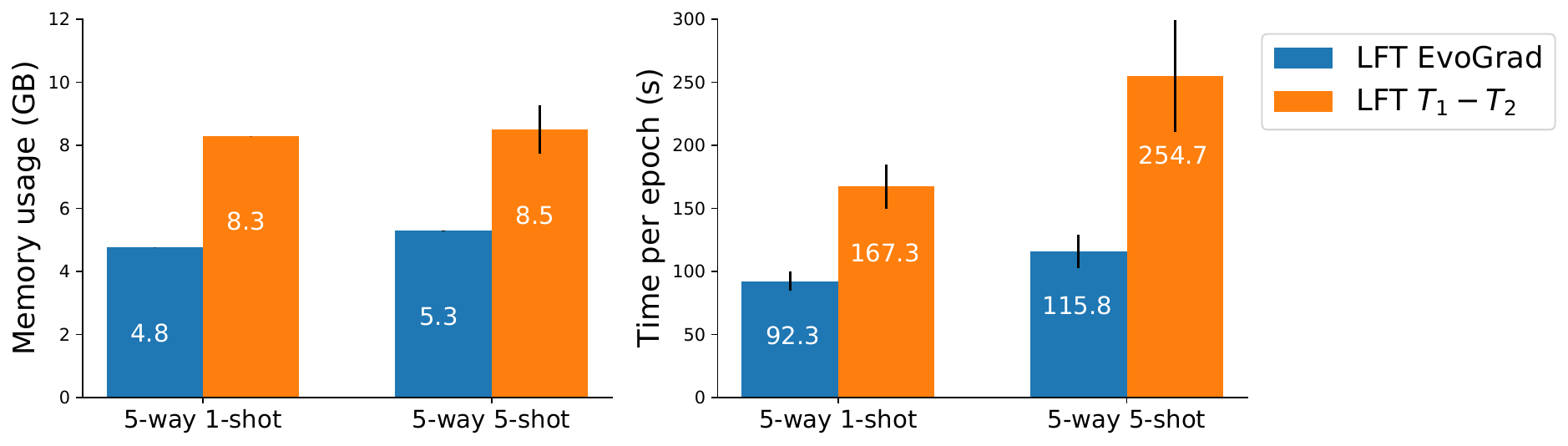}
  \caption{Cross-domain few-shot learning with LFT \citep{Tseng2020Cross-domainTransformation}: analysis of memory and time efficiency of \algName{} vs standard second-order $T_1-T_2$ approach. Mean and standard deviation reported across experiments with different test datasets. EvoGrad is significantly more efficient in terms of both memory usage and time per epoch.}
  \label{fig:cdfsleff}
\end{figure}

Table \ref{tab:cdfsl5w1s} shows the baseline performance of vanilla unregularised ResNet (-), manually tuned FT layers (FT), FT layers meta-learned by second-order gradient (LFT) and by EvoGrad. The results show that EvoGrad matches the accuracy of the original LFT approach, leading to clear accuracy improvements over training with no feature-wise transformation or training with fixed feature-wise parameters selected manually. At the same time EvoGrad is significantly more efficient in terms of the memory and time costs as shown in Figure \ref{fig:cdfsleff}. The memory improvements from EvoGrad allow us to scale the base feature extractor to ResNet34 within the standard 12GB GPU. The original LFT with its $T_1-T_2$ style second-order algorithm cannot be extended in the available memory if we keep the same settings of the few-shot learning tasks. Thus, we are able to improve state-of-the-art accuracy on both 5-way 1 and 5-shot tasks. For ResNet34, we include baselines without any feature-wise transformation and with manually chosen feature-wise transformation to confirm the benefit of meta-learning.

\begin{table}[h!]
\caption{RelationNet test accuracies (\%) and 95\% confidence intervals across test tasks on various unseen datasets. LFT EvoGrad can scale to ResNet34 on all tasks within 12GB GPU memory, while vanilla second-order LFT $T_1-T_2$ cannot. We also report the results of our own rerun of the LFT approach using the official code -- denoted as \textit{our run}. EvoGrad can clearly match the accuracies obtained by the original approach that uses $T_1-T_2$.}
\label{tab:cdfsl5w1s}
\centering
\resizebox{1.0\textwidth}{!}{
\begin{tabular}{cllcccc}
\toprule
 & Model & Approach & CUB & Cars & Places & Plantae \\
\midrule
\multirow{8}{*}{\rotatebox[origin=c]{90}{5-way 1-shot}}&\multirow{5}{*}{ResNet10}& - & 44.33 $\pm$ 0.59 & 29.53 $\pm$ 0.45 & 47.76 $\pm$ 0.63 & 33.76 $\pm$ 0.52 \\
&& FT & 44.67 $\pm$ 0.58 & 30.38 $\pm$ 0.47 & 48.40 $\pm$ 0.64 & 35.40 $\pm$ 0.53 \\
&& LFT $T_1-T_2$ & 48.38 $\pm$ 0.63 & 32.21 $\pm$ 0.51 & 50.74 $\pm$ 0.66 & 35.00 $\pm$ 0.52 \\
&& LFT $T_1-T_2$ (our run) & 46.03 $\pm$ 0.60 & 31.50 $\pm$ 0.49 & 49.29 $\pm$ 0.65 & 36.34 $\pm$ 0.59 \\
&& LFT EvoGrad & 47.39 $\pm$ 0.61 & 32.51 $\pm$ 0.56 & 50.70 $\pm$ 0.66 & 36.00 $\pm$ 0.56 \\
\cline{2-7}
& \multirow{3}{*}{ResNet34} & - & 45.61 $\pm$ 0.59 & 29.54 $\pm$ 0.46 & 48.87 $\pm$ 0.65 & 35.03 $\pm$ 0.54 \\
&& FT & 45.15 $\pm$ 0.59 & 30.28 $\pm$ 0.44 & 49.96 $\pm$ 0.66 & 35.69 $\pm$ 0.54 \\
&& LFT EvoGrad & 45.97 $\pm$ 0.60 & 33.21 $\pm$ 0.54 & 50.76 $\pm$ 0.67 & 38.23 $\pm$ 0.58 \\
\bottomrule
\multirow{9}{*}{\rotatebox[origin=c]{90}{5-way 5-shot}} & \multirow{5}{*}{ResNet10} & - & 62.13 $\pm$ 0.74 & 40.64 $\pm$ 0.54 & 64.34 $\pm$ 0.57 & 46.29 $\pm$ 0.56 \\
&& FT & 63.64 $\pm$ 0.77 & 42.24 $\pm$ 0.57 & 65.42 $\pm$ 0.58 & 47.81 $\pm$ 0.51 \\
&& LFT $T_1-T_2$ & 64.99 $\pm$ 0.54 & 43.44 $\pm$ 0.59 & 67.35 $\pm$ 0.54 & 50.39 $\pm$ 0.52 \\
&& LFT $T_1-T_2$ (our run) & 65.94 $\pm$ 0.56 & 43.88 $\pm$ 0.56 & 65.57 $\pm$ 0.57 & 51.43 $\pm$ 0.55 \\
&& LFT EvoGrad & 64.63 $\pm$ 0.56 & 42.64 $\pm$ 0.58 & 66.54 $\pm$ 0.57 & 52.92 $\pm$ 0.57 \\
\cline{2-7}
&\multirow{3}{*}{ResNet34} & - & 63.33 $\pm$ 0.59 & 40.50 $\pm$ 0.55 & 64.94 $\pm$ 0.56 & 50.20 $\pm$ 0.55 \\
&& FT & 62.48 $\pm$ 0.56 & 41.06 $\pm$ 0.52 & 64.39 $\pm$ 0.57 & 50.08 $\pm$ 0.55 \\
&& LFT EvoGrad & 66.40 $\pm$ 0.56 & 44.25 $\pm$ 0.55 & 67.23 $\pm$ 0.56 & 52.47 $\pm$ 0.56 \\
\bottomrule
\end{tabular}}
\end{table}

\subsection{Label noise with Meta-Weight-Net}
We consider a further highly practical real problem where online meta-learning has led to significant improvements -- learning from noisy labelled data. The Meta-Weight-Net framework trains an auxiliary neural network that performs instance-wise loss re-weighting on the training set \citep{Shu2019Meta-Weight-Net:Weighting}. The base model is updated using the sum of weighted instance-wise losses for noisy data, while the Meta-Weight-Net itself is updated by evaluating the updated model on clean validation data and by backpropagating through the model update. We use the official implementation of the approach \citep{Shu2019Meta-Weight-Net:Weighting} and follow the same experimental settings, using $K=2$ for \algName{}.

\begin{figure}[h!]
  \centering
  \includegraphics[width=0.9\textwidth]{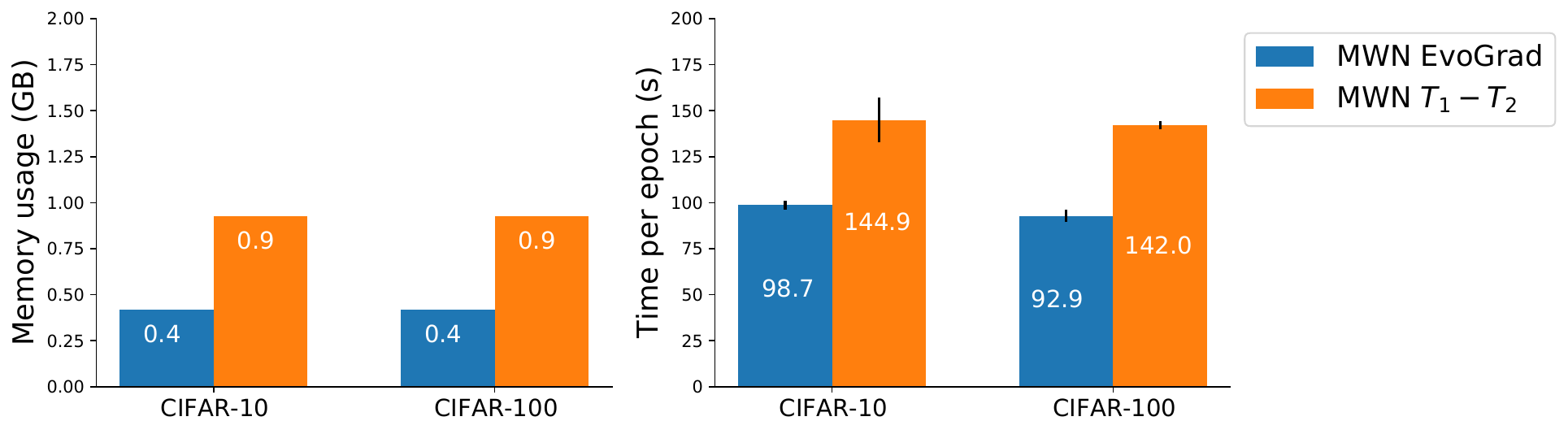}
  \caption{Analysis of memory and time cost of MWN \algName{} vs the original second-order MWN, showing significant efficiency improvements by EvoGrad. Mean and standard deviation is reported across 5 repetitions of 40\% label noise problem.}
  \label{fig:mwnAnalysis}
\end{figure}

Our results in Table \ref{tab:mwnresults} confirm we replicate the benefits of training with Meta-Weight-Net, clearly surpassing the accuracy of the baseline when there is label noise. We also note that EvoGrad can improve the accuracy over the $T_1-T_2$-based approach because the two approaches are distinct and provide different estimates of the true hypergradient. Figure \ref{fig:mwnAnalysis} shows that our method leads to significant improvements in memory and time costs (over half of the memory is saved and the runtime is improved by about a third). 

\begin{table}[h!]
\caption{Test accuracies (\%) for Meta-Weight-Net label noise experiments with ResNet-32 -- means and standard deviations across 5 repetitions for the original second-order algorithm vs \algName{}. \algName{} is able to match or even exceed the accuracies obtained by the original MWN approach.}
\label{tab:mwnresults}
\centering
\resizebox{1.0\textwidth}{!}{
\begin{tabular}{lccccc}
\toprule
Dataset & Noise rate & Baseline & MWN $T_1-T_2$ & MWN $T_1-T_2$ (our run) & MWN EvoGrad  \\
\midrule
 & 0\% & 92.89 $\pm$ 0.32 & 92.04 $\pm$ 0.15 & 91.10 $\pm$ 0.19 & 92.02 $\pm$ 0.31 \\
CIFAR-10 & 20\% & 76.83 $\pm$ 2.30 & 90.33 $\pm$ 0.61 & 89.31 $\pm$ 0.40 & 89.86 $\pm$ 0.64 \\
 & 40\% & 70.77 $\pm$ 2.31 & 87.54 $\pm$ 0.23 & 85.90 $\pm$ 0.45 & 87.74 $\pm$ 0.54 \\
\midrule
 & 0\% & 70.50 $\pm$ 0.12 & 70.11 $\pm$ 0.33 & 68.42 $\pm$ 0.36 & 69.16 $\pm$ 0.49\\
CIFAR-100 & 20\% & 50.86 $\pm$ 0.27 & 64.22 $\pm$ 0.28 & 63.43 $\pm$ 0.43 & 64.05 $\pm$ 0.63 \\
 & 40\% & 43.01 $\pm$ 1.16 & 58.64 $\pm$ 0.47 & 56.54 $\pm$ 0.90 & 57.44 $\pm$ 1.25 \\
\bottomrule
\end{tabular}}
\end{table}

\subsection{Low-resource cross-lingual learning with MetaXL}
The previous two real applications of meta-learning considered computer vision problems. To highlight EvoGrad is a general method that can make an impact in any domain, we also demonstrate its benefits on a meta-learning application from NLP. More specifically, we use EvoGrad for MetaXL \citep{Xia2021MetaXL:Learning}, which meta-learns meta representation transformation to better transfer from source languages to low-resource target languages.

\begin{figure}[h!]
  \centering
  \includegraphics[width=0.9\textwidth]{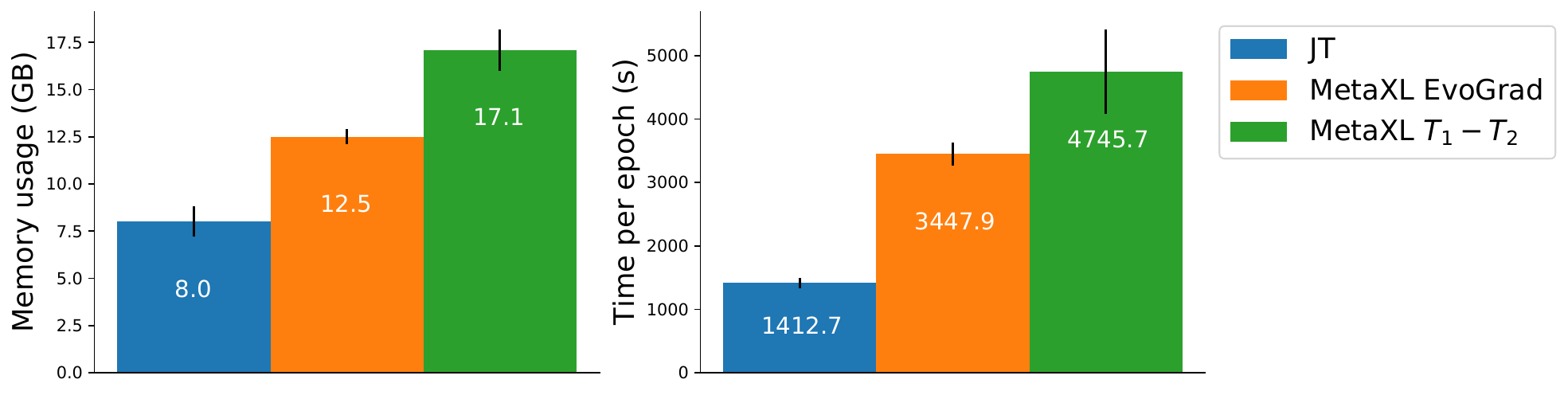}
  \caption{Analysis of memory and time cost of MetaXL \algName{} vs the original second-order MetaXL, in the context of a simple joint-training (JT) baseline. \algName{} consumes significantly less memory than $T_1-T_2$ and is faster. Mean and standard deviation is calculated over the 8 different target languages.}
  \label{fig:metaxlAnalysis}
\end{figure}

We have selected the named entity recognition (NER) task with English source language (WikiAnn dataset \citep{Pan2017Cross-lingualLanguages}), which is one of the key experiments in the MetaXL paper \citep{Xia2021MetaXL:Learning}. Table \ref{tab:metaxlf1score} shows EvoGrad matches and in fact surpasses the average test F1 score of MetaXL with the original $T_1-T_2$ meta-learning method. Figure \ref{fig:metaxlAnalysis} shows EvoGrad significantly improves both memory and time consumption compared to MetaXL $T_1-T_2$. Overall these results confirm EvoGrad is suitable for meta-learning in various domains, including both computer vision and NLP.

\begin{table}[h!]
\caption{Test F1 score in \% for named entity recognition task. English source language. The first two rows are taken from the MetaXL paper, while our own runs are in the following rows. EvoGrad clearly matches and even surpasses the performance of $T_1-T_2$ baseline. Joint-training (JT) represents a simple non-meta-learning baseline approach.}
\label{tab:metaxlf1score}
\centering
\resizebox{1.0\textwidth}{!}{
\begin{tabular}{lccccccccc}
\toprule
Method                   & qu    & cdo   & ilo   & xmf   & mhr   & mi    & tk    & gn    & Average \\
\midrule
JT                       & 66.10 & 55.83 & 80.77 & 69.32 & 71.11 & 82.29 & 61.61 & 65.44 & 69.06   \\
MetaXL $T_1-T_2$         & 68.67 & 55.97 & 77.57 & 73.73 & 68.16 & 88.56 & 66.99 & 69.37 & 71.13   \\
JT (our run)             & 59.75 & 49.19 & 79.43 & 68.85 & 68.42 & 89.94 & 61.90 & 69.44 & 68.37   \\
MetaXL $T_1-T_2$ (our run)   & 65.29 & 56.33 & 76.50 & 67.24 & 71.17 & 89.41 & 66.67 & 64.11 & 69.59   \\
MetaXL EvoGrad & 71.00 & 57.02 & 85.99 & 70.40 & 65.45 & 88.12 & 66.97 & 70.91 & 71.98 \\
\bottomrule
\end{tabular}}
\end{table}

\subsection{Scalability analysis}
We use the Meta-Weight-Net benchmark to study how the number of model parameters affects the memory usage and training time of EvoGrad, comparing it to the standard second-order $T_1-T_2$ approach. We vary model size by changing the number of filters in the original ResNet32 model, multiplying the filter number $\times1,\dots,\times5$.
The smallest model had around 0.5M parameters and the largest one around 11M parameters. 

\begin{figure}[h!]
  \centering
  \includegraphics[width=\textwidth]{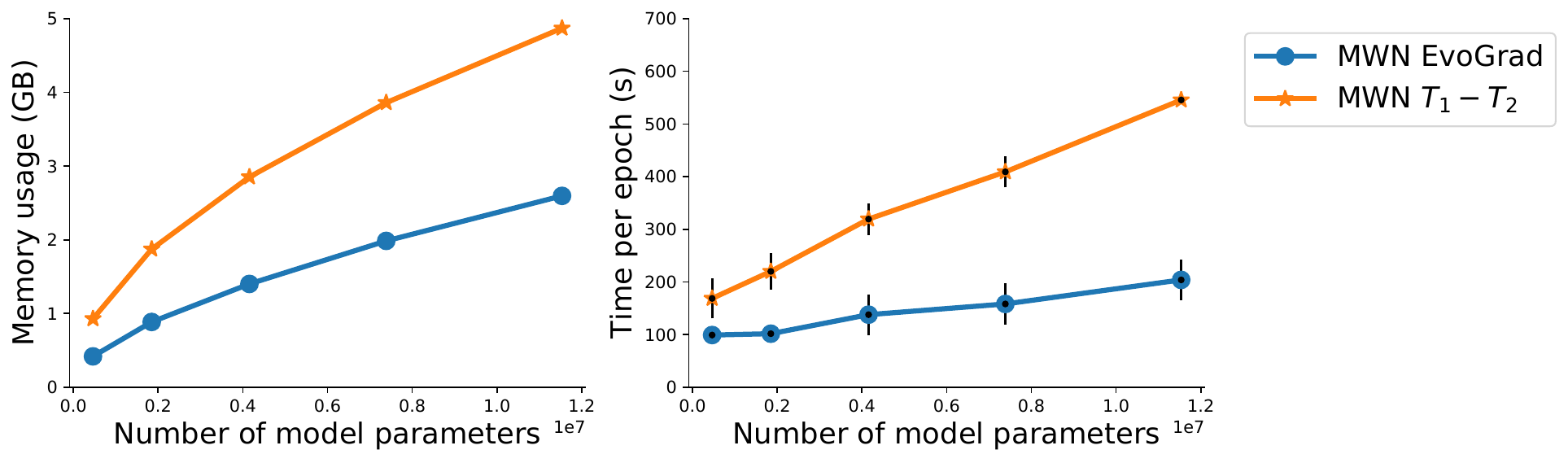}
  \caption{Memory and time scaling of MWN \algName{} vs original second-order Meta-Weight-Net. Efficiency margins of EvoGrad are larger for larger models.}
  \label{fig:mwnAnalysisScaled}
\end{figure}

The results in Figure \ref{fig:mwnAnalysisScaled} show our \algName{} leads to significantly lower training time and memory usage, and that the margin over the standard second-order optimizer grows as the model becomes larger. Further, we have analysed the impact of modifying the number of hyperparameters -- from 300 up to 30,000. The impact on memory and time was negligible, and both remained roughly constant, which is caused by the main model being significantly larger. It is also because of the fact that reverse-mode differentiation costs scale with the number of model parameters rather than hyperparameters \citep{Micaelli2020Non-greedyHorizons} -- recall that backpropagation is the main driver of memory and time costs \citep{Rajeswaran2019Meta-learningGradients}. Moreover, we have done experiments that varied the number of model copies in EvoGrad. The results showed the training time per epoch increased slightly, while the memory costs remained similar.

\section{Discussion and limitations}
Similar to many other gradient-based meta-learning methods, our method is greedy as it considers only the current state of the model when updating the hyperparameters -- rather than the whole training process. However, this greediness allows the method to be used in larger-scale settings where we train the hyperparameters and the model jointly.
Further, our method approximates the hypergradient stochastically. While results were good for the suite of problems considered here using only $K=2$, the gradient estimates may be too noisy in other applications. This could lead to poor outcomes which could be a problem in socially important applications. Alternatively, it may necessitate using a larger model population (Figure~\ref{fig:toy1dgrads}). While as we observed in Section~\ref{sec:project} the candidate models can be trivially parallelized to scale population size, this still imposes a larger energy cost \cite{Schwartz2019GreenAI}. Another limitation is that similarly to IFT-based estimators \cite{Lorraine2020OptimizingDifferentiation}, EvoGrad is not suitable for optimizing learner hyperparameters such as learning rate. Currently we have used the simplest possible evolutionary update in the inner loop, and upgrading EvoGrad to a state-of-the-art evolutionary strategy may lead to better gradient estimates and improve results further.

\section{Conclusions}
We have proposed a new efficient method for meta-learning that allows us to scale gradient-based meta-learning to bigger models and problems. We have evaluated the method on a variety of problems, most notably meta-learning feature-wise transformation layers, training with noisy labels using Meta-Weight-Net, and meta-learning meta representation transformation for low-resource cross-lingual learning. In all cases we have shown significant time and memory efficiency improvements, while achieving similar or better performance compared to the existing meta-learning methods.

\begin{ack}
This work was supported in part by the EPSRC Centre for Doctoral Training in Data Science, funded by the UK Engineering and Physical Sciences Research Council (grant EP/L016427/1) and the University of Edinburgh.
\end{ack}

\bibliographystyle{apalike}
\bibliography{references,moreref}

\begin{thebibliography}{}

\bibitem[Antoniou et~al., 2019]{Antoniou2019HowMAML}
Antoniou, A., Edwards, H., and Storkey, A. (2019).
\newblock {How to train your MAML}.
\newblock In {\em ICLR}.

\bibitem[Balaji et~al., 2018]{Balaji2018MetaReg:Meta-regularization}
Balaji, Y., Sankaranarayanan, S., and Chellappa, R. (2018).
\newblock {MetaReg: towards domain generalization using meta-regularization}.
\newblock In {\em NeurIPS}.

\bibitem[Baydin et~al., 2018]{Baydin2018AutomaticSurvey}
Baydin, A.~G., Pearlmutter, B.~A., and Siskind, J.~M. (2018).
\newblock {Automatic differentiation in machine learning: a survey}.
\newblock {\em Journal of Machine Learning Research}, 18:1--43.

\bibitem[Bengio, 2000]{Bengio2000Gradient-basedHyperparameters}
Bengio, Y. (2000).
\newblock {Gradient-based optimization of hyperparameters}.
\newblock {\em Neural Computation}, 12(8):1889--1900.

\bibitem[Bohdal et~al., 2020]{Bohdal2020FlexibleImages}
Bohdal, O., Yang, Y., and Hospedales, T. (2020).
\newblock {Flexible dataset distillation: learn labels instead of images}.
\newblock In {\em NeurIPS MetaLearn 2020 workshop}.

\bibitem[Chen et~al., 2019]{Chen2019AClassification}
Chen, W.-Y., Liu, Y.-C., Kira, Z., Tech, G., Wang, Y.-C.~F., Huang, J.-B., and
  Tech, V. (2019).
\newblock {A closer look at few-shot classification}.
\newblock In {\em ICLR}.

\bibitem[Cubuk et~al., 2019]{cubuk2019autoAug}
Cubuk, E.~D., Zoph, B., Man{\'{e}}, D., Vasudevan, V., and Le, Q.~V. (2019).
\newblock Autoaugment: learning augmentation policies from data.
\newblock {\em CVPR}.

\bibitem[Finn et~al., 2017]{Finn2017Model-agnosticNetworks}
Finn, C., Abbeel, P., and Levine, S. (2017).
\newblock {Model-agnostic meta-learning for fast adaptation of deep networks}.
\newblock In {\em ICML}.

\bibitem[Grefenstette et~al., 2019]{Grefenstette2019GeneralizedMeta-Learning}
Grefenstette, E., Amos, B., Yarats, D., Htut, P.~M., Molchanov, A., Meier, F.,
  Kiela, D., Cho, K., and Chintala, S. (2019).
\newblock {Generalized inner loop meta-learning}.
\newblock In {\em arXiv}.

\bibitem[Griewank, 1993]{Griewank1993SomeHessians}
Griewank, A. (1993).
\newblock {Some bounds on the complexity of gradients, Jacobians, and
  Hessians}.
\newblock In {\em Complexity in Numerical Optimization}, pages 128--162.

\bibitem[He et~al., 2015]{He2015DeepRecognition}
He, K., Zhang, X., Ren, S., and Sun, J. (2015).
\newblock {Deep residual learning for image recognition}.
\newblock In {\em CVPR}.

\bibitem[Horn et~al., 2018]{Horn2018TheDataset}
Horn, G.~V., Aodha, O.~M., Song, Y., Cui, Y., Sun, C., Shepard, A., Adam, H.,
  Perona, P., and Belongie, S. (2018).
\newblock {The iNaturalist species classification and detection dataset}.
\newblock In {\em CVPR}.

\bibitem[Hospedales et~al., 2021]{Hospedales2021Meta-learningSurvey}
Hospedales, T.~M., Antoniou, A., Micaelli, P., and Storkey, A.~J. (2021).
\newblock {Meta-learning in neural networks: a survey}.
\newblock {\em IEEE Transactions on Pattern Analysis and Machine Intelligence},
  PP:1--1.

\bibitem[Jaderberg et~al., 2019]{jaderberg2019popRL}
Jaderberg, M., Czarnecki, W.~M., Dunning, I., Marris, L., Lever, G.,
  Casta{\~n}eda, A.~G., Beattie, C., Rabinowitz, N.~C., Morcos, A.~S.,
  Ruderman, A., Sonnerat, N., Green, T., Deason, L., Leibo, J.~Z., Silver, D.,
  Hassabis, D., Kavukcuoglu, K., and Graepel, T. (2019).
\newblock Human-level performance in 3d multiplayer games with population-based
  reinforcement learning.
\newblock {\em Science}, 364(6443):859--865.

\bibitem[Kingma and Ba, 2015]{Kingma2015Adam:Optimization}
Kingma, D.~P. and Ba, J. (2015).
\newblock {Adam: a method for stochastic optimization}.
\newblock In {\em ICLR}.

\bibitem[Krause et~al., 2013]{Krause20133DCategorization}
Krause, J., Stark, M., Deng, J., and Fei-Fei, L. (2013).
\newblock {3D object representations for fine-grained categorization}.
\newblock In {\em ICCV Workshops}.

\bibitem[Krizhevsky, 2009]{Krizhevsky2009LearningImages}
Krizhevsky, A. (2009).
\newblock {Learning multiple layers of features from tiny images}.
\newblock Technical report.

\bibitem[Larsen et~al., 1996]{Larsen1996DesignSet}
Larsen, J., Hansen, L.~K., Svarer, C., and Ohlsson, M. (1996).
\newblock {Design and regularization of neural networks: the optimal use of a
  validation set}.
\newblock In {\em Neural Networks for Signal Processing - Proceedings of the
  IEEE Workshop}, pages 62--71. IEEE.

\bibitem[LeCun et~al., 1998]{LeCun1998MNISTDatabase}
LeCun, Y., Cortes, C., and Burges, C. (1998).
\newblock {MNIST handwritten digit database}.

\bibitem[LeCun et~al., 1989]{LeCun1989HandwrittenLearning}
LeCun, Y., Jackel, L.~D., Boser, B., Denker, J.~S., Graf, H.~P., Guyon, I.,
  Henderson, D., Howard, R.~E., and Hubbard, W. (1989).
\newblock {Handwritten digit recognition: applications of neural network chips
  and automatic learning}.
\newblock {\em IEEE Communications Magazine}, 27(11):41--46.

\bibitem[Li et~al., 2019]{Li2019Feature-criticGeneralization}
Li, Y., Yang, Y., Zhou, W., and Hospedales, T.~M. (2019).
\newblock {Feature-critic networks for heterogeneous domain generalization}.
\newblock In {\em ICML}.

\bibitem[Li et~al., 2017]{Li2017Meta-SGD:Learning}
Li, Z., Zhou, F., Chen, F., and Li, H. (2017).
\newblock {Meta-SGD: learning to learn quickly for few-shot learning}.
\newblock In {\em arXiv}. arXiv.

\bibitem[Liu et~al., 2019a]{liu2019darts}
Liu, H., Simonyan, K., and Yang, Y. (2019a).
\newblock {DARTS}: differentiable architecture search.
\newblock In {\em ICLR}.

\bibitem[Liu et~al., 2019b]{Liu2019Self-SupervisedLearning}
Liu, S., Davison, A.~J., and Johns, E. (2019b).
\newblock {Self-supervised generalisation with meta auxiliary learning}.
\newblock In {\em NeurIPS}.

\bibitem[Lorraine and Duvenaud, 2018]{Lorraine2018StochasticHypernetworks}
Lorraine, J. and Duvenaud, D. (2018).
\newblock {Stochastic hyperparameter optimization through hypernetworks}.
\newblock In {\em arXiv}.

\bibitem[Lorraine et~al., 2020]{Lorraine2020OptimizingDifferentiation}
Lorraine, J., Vicol, P., and Duvenaud, D. (2020).
\newblock {Optimizing millions of hyperparameters by implicit differentiation}.
\newblock In {\em AISTATS}.

\bibitem[Luketina et~al., 2016]{Luketina2016ScalableHyperparameters}
Luketina, J., Berglund, M., Klaus~Greff, A., and Raiko, T. (2016).
\newblock {Scalable gradient-based tuning of continuous regularization
  hyperparameters}.
\newblock In {\em ICML}.

\bibitem[Maclaurin et~al., 2015]{Maclaurin2015Gradient-basedLearning}
Maclaurin, D., Duvenaud, D., and Adams, R.~P. (2015).
\newblock {Gradient-based hyperparameter optimization through reversible
  learning}.
\newblock In {\em ICML}.

\bibitem[Micaelli and Storkey, 2019]{Micaelli2019Zero-shotMatching}
Micaelli, P. and Storkey, A. (2019).
\newblock {Zero-shot knowledge transfer via adversarial belief matching}.
\newblock In {\em NeurIPS}.

\bibitem[Micaelli and Storkey, 2020]{Micaelli2020Non-greedyHorizons}
Micaelli, P. and Storkey, A. (2020).
\newblock {Non-greedy gradient-based hyperparameter optimization over long
  horizons}.
\newblock In {\em arXiv}.

\bibitem[Nichol et~al., 2018]{Nichol2018OnAlgorithms}
Nichol, A., Achiam, J., and Schulman, J. (2018).
\newblock {On first-order meta-learning algorithms}.
\newblock In {\em arXiv}.

\bibitem[Pan et~al., 2017]{Pan2017Cross-lingualLanguages}
Pan, X., Zhang, B., May, J., Nothman, J., Knight, K., and Ji, H. (2017).
\newblock {Cross-lingual name tagging and linking for 282 languages}.
\newblock In {\em ACL}.

\bibitem[Rajeswaran et~al., 2019]{Rajeswaran2019Meta-learningGradients}
Rajeswaran, A., Finn, C., Kakade, S., and Levine, S. (2019).
\newblock {Meta-learning with implicit gradients}.
\newblock In {\em NeurIPS}.

\bibitem[Ravi and Larochelle, 2017]{Ravi2017OptimizationLearning}
Ravi, S. and Larochelle, H. (2017).
\newblock {Optimization as a model for few-shot learning}.
\newblock In {\em ICLR}.

\bibitem[Ren et~al., 2018]{Ren2018LearningLearning}
Ren, M., Zeng, W., Yang, B., and Urtasun, R. (2018).
\newblock {Learning to reweight examples for robust deep learning}.
\newblock In {\em ICML}, volume~10.

\bibitem[Salimans et~al., 2017]{salimans2017evolution}
Salimans, T., Ho, J., Chen, X., Sidor, S., and Sutskever, I. (2017).
\newblock Evolution strategies as a scalable alternative to reinforcement
  learning.
\newblock In {\em arXiv}.

\bibitem[Schwartz et~al., 2019]{Schwartz2019GreenAI}
Schwartz, R., Dodge, J., Smith, N.~A., and Etzioni, O. (2019).
\newblock {Green AI}.
\newblock In {\em arXiv}.

\bibitem[Shaban et~al., 2019]{Shaban2019TruncatedOptimization}
Shaban, A., Cheng, C.-A., Hatch, N., and Boots, B. (2019).
\newblock {Truncated back-propagation for bilevel optimization}.
\newblock In {\em AISTATS}.

\bibitem[Shan et~al., 2020]{Shan2020Meta-Neighborhoods}
Shan, S., Li, Y., and Oliva, J.~B. (2020).
\newblock {Meta-Neighborhoods}.
\newblock In {\em NeurIPS}.

\bibitem[Shu et~al., 2019]{Shu2019Meta-Weight-Net:Weighting}
Shu, J., Xie, Q., Yi, L., Zhao, Q., Zhou, S., Xu, Z., and Meng, D. (2019).
\newblock {Meta-Weight-Net: learning an explicit mapping for sample weighting}.
\newblock In {\em NeurIPS}.

\bibitem[Sung et~al., 2018]{Sung2018LearningLearning}
Sung, F., Yang, Y., Zhang, L., Xiang, T., Torr, P.~H., and Hospedales, T.~M.
  (2018).
\newblock {Learning to compare: relation network for few-shot learning}.
\newblock In {\em CVPR}.

\bibitem[Tseng et~al., 2020]{Tseng2020Cross-domainTransformation}
Tseng, H.-Y., Lee, H.-Y., Huang, J.-B., and Yang, M.-H. (2020).
\newblock {Cross-domain few-shot classification via learned feature-wise
  transformation}.
\newblock In {\em ICLR}.

\bibitem[Wang et~al., 2018]{Wang2018DatasetDistillation}
Wang, T., Zhu, J.-Y., Torralba, A., and Efros, A.~A. (2018).
\newblock {Dataset distillation}.
\newblock In {\em arXiv}.

\bibitem[Welinder et~al., 2010]{Welinder2010Caltech-UCSD200}
Welinder, P., Branson, S., Mita, T., Wah, C., Schroff, F., Belongie, S., and
  Perona, P. (2010).
\newblock {Caltech-UCSD Birds 200}.
\newblock Technical report, California Institute of Technology.

\bibitem[Xia et~al., 2021]{Xia2021MetaXL:Learning}
Xia, M., Zheng, G., Mukherjee, S., Shokouhi, M., Neubig, G., and Awadallah,
  A.~H. (2021).
\newblock {MetaXL: meta representation transformation for low-resource
  cross-lingual learning}.
\newblock In {\em NAACL}.

\bibitem[Zhou et~al., 2018]{Zhou2018Places:Recognition}
Zhou, B., Lapedriza, A., Khosla, A., Oliva, A., and Torralba, A. (2018).
\newblock {Places: a 10 million image database for scene recognition}.
\newblock {\em IEEE Transactions on Pattern Analysis and Machine Intelligence},
  40(6):1452--1464.

\bibitem[Zhou et~al., 2019]{Zhou2019EfficientUpdate}
Zhou, P., Yuan, X.-T., Xu, H., Yan, S., and Feng, J. (2019).
\newblock {Efficient meta learning via minibatch proximal update}.
\newblock In {\em NeurIPS}.

\bibitem[Zoph and Le, 2017]{zoph2017neuralArchitectureRL}
Zoph, B. and Le, Q.~V. (2017).
\newblock Neural architecture search with reinforcement learning.
\newblock In {\em ICLR}.

\end{thebibliography}
\clearpage
\appendix
\section{Code illustration}
\subsection{EvoGrad code}
EvoGrad update is simple to implement if we use \textit{higher} library \citep{Grefenstette2019GeneralizedMeta-Learning} for the perturbed parameters of different model copies. We show only the part that is relevant to the meta-update.
\begin{lstlisting}[language=Python, caption=EvoGrad code example.]
model_parameter = [i.detach() for i in get_func_params(model)]
theta_list = [[j + sigma * torch.sign(torch.randn_like(j))
              for j in model_parameter] for i in range(n_model_candidates)]
pred_list = [model_patched(feature_transformer(inputs), params=theta)
             for theta in theta_list]
loss_list = [criterion(pred, targets) for pred in pred_list]
weights = torch.softmax(-torch.stack(loss_list) / temperature, 0)
theta_updated = [sum(map(mul, theta, weights))
                 for theta in zip(*theta_list)]
preds_meta = model_patched(inputs_meta, params=theta_updated)
loss_meta = criterion(preds_meta, targets_meta)

meta_opt.zero_grad()
loss_meta.backward()
meta_opt.step()
\end{lstlisting}

\subsection{T1--T2 code -- for comparison}
For comparison with EvoGrad, we also show how online $T1-T2$-style meta-learning is often implemented using so-called \textit{fast weights}. This approach has been, for example, used in \citep{Chen2019AClassification,Tseng2020Cross-domainTransformation}. The meta-update itself is concise, but it requires us to implement layers that support fast weights, which is a significantly longer part.

\begin{lstlisting}[language=Python, caption=$T1-T2$ code example.]
preds = model(feature_transformer(inputs))
loss = criterion(preds, targets)
optimizer.zero_grad()
grads = torch.autograd.grad(loss, model.parameters(), create_graph=True)
for k, weight in enumerate(model.parameters()):
    weight.fast = weight - meta_lr * grads[k]

preds_meta = model(inputs_meta)
loss_meta = criterion(preds_meta, targets_meta)

meta_opt.zero_grad()
loss_meta.backward()
meta_opt.step()
\end{lstlisting}

We also show the definition of a linear layer that supports fast weights:
\begin{lstlisting}[language=Python, caption=Code example for a linear layer that supports fast weights.]
class Linear_fw(nn.Linear):
    def __init__(self, in_features, out_features, bias=True):
        super(Linear_fw, self).__init__(in_features, out_features,
                                        bias=bias)
        self.weight.fast = None
        self.bias.fast = None

    def forward(self, x):
        if self.weight.fast is not None and self.bias.fast is not None:
            out = F.linear(x, self.weight.fast, self.bias.fast)
        else:
            out = super(Linear_fw, self).forward(x)
        return out
\end{lstlisting}
Normally we would use simple \texttt{nn.Linear(in\_features, out\_features)}.

\section{How to select EvoGrad hyperparameters}
EvoGrad as an algorithm has a few hyperparameters common to most evolutionary approaches: perturbation value $\sigma$, temperature $\tau$ and the number of model copies $K$. In practice we use only 2 models as it is enough and improves the efficiency. The other hyperparameter values can be selected relatively easily by looking at the training loss of the unperturbed model and the training loss of the perturbed models. The losses should be similar to each other, but not the same -- we want to make sure the perturbed weights can still be successfully used. We have found that in practice value $\sigma=0.001$ is reasonable. Once we have selected the value of $\sigma$, we can select the value of temperature $\tau$ which leads to reasonably different weights for the two (or more) model copies. In practice we have found $\tau=0.05$ to be a value which leads to suitable weights. For example, 0.48 and 0.52 for two model copies could be considered reasonable, while 0.5001 and 0.4999 would be too similar. Note that in the special case of a 1-dimensional toy problem, suitable EvoGrad hyperparameters are different than what is useful for practical problems.

\section{Additional details}
We include an algorithmic description of the details as well as additional description of the experimental settings for all five problems that we discuss in the paper.

\subsection{Illustration using a 1-dimensional problem}
We provide more detailed descriptions of how we perform both analyses. In the first analysis, we calculate the EvoGrad hypergradient estimate for 100 values of $\lambda$ between 0 and 2, starting with 0.1 and ending with 2.0. In each case we perform 100 repetitions to obtain an estimate of the mean and standard deviation of the hypergradient, considering the stochastic nature of EvoGrad. Given a value of $\lambda$, the process of EvoGrad estimate can be summarized using Algorithm \ref{alg:evograd1dest}. As a reminder, we use training loss function $f_T(x, \lambda)=(x-1)^2+\lambda \|x\|_2^2$ that includes a meta-parameter $\lambda$ and validation loss function $f_V(x)=(x-0.5)^2$ that does not include the meta-parameter. The value of temperature is 0.5 and the number of model candidates varies between 2, 10 and 100.

\begin{algorithm}[h]
   \caption{EvoGrad hypergradient estimate for the 1D problem}
   \label{alg:evograd1dest}
\begin{algorithmic}[1]
   \STATE {\bfseries Input:} $\lambda$: target hyperparameter; $K$: number of model candidates; $\tau$: temperature; $f_T, f_V$: training and validation loss functions 
   \STATE {\bfseries Output:} $g$: hypergradient estimate
   \STATE Sample $x \sim \mathcal{N}(0, 1)$
   \STATE Sample $K$ noise parameters $\epsilon_k\sim \mathcal{N}(0, 1)$ and use them to create $x_k=x+\epsilon_k$
   \STATE Calculate losses $\ell_k=f_T(x_k, \lambda)$ for $k$ between $1$ and $K$
   \STATE Calculate weights $w_1, w_2, \dots, w_K = \operatorname{softmax}([-\ell_1, -\ell_2, \dots, -\ell_K]/\tau)$
   \STATE Calculate $x^* = w_1 x_1 + w_2 x_2 + \dots + w_K x_K$
   \STATE Calculate $\ell_V=f_V(x^*)$
   \STATE Calculate hypergradient $g=\frac{\partial \ell_V}{\partial \lambda}$ by backpropagating through $x^*$ computation
\end{algorithmic}
\end{algorithm}

The second analysis evaluates the trajectories that values of $x, \lambda$ take if we update them with SGD with the hypergradient estimated by EvoGrad compared to the ground-truth. We can summarize the process using Algorithm \ref{alg:evograd1dtraj}. When using the ground-truth hypergradient, we simply replace lines 6 to 10 by directly updating the value of $\lambda$ using the closed-form formula for the hypergradient: $g(\lambda)=(\lambda-1)/(\lambda+1)^3$. We use 5 steps, learning rate of 0.1 and temperature 0.5.

\begin{algorithm}[h]
   \caption{Training with EvoGrad -- 1D problem}
   \label{alg:evograd1dtraj}
\begin{algorithmic}[1]
   \STATE {\bfseries Input:} $x_0, \lambda_0$: initial values of $x, \lambda$; $N$: number of steps; $\alpha$: learning rate; $K$: number of model candidates; $\tau$: temperature; $f_T, f_V$: training and validation loss functions 
   \STATE {\bfseries Output:} Optimized values of $x, \lambda$
   \STATE Initialize $x=x_0$ and $\lambda=\lambda_0$
   \FOR{$i$ between $1$ and $N$}
   \STATE Update $x \leftarrow x - \alpha \frac{\partial f_T(x, \lambda)}{\partial x}$
   \STATE Sample $K$ noise parameters $\epsilon_k\sim \mathcal{N}(0, 1)$ and use them to create $x_k=x+\epsilon_k$
   \STATE Calculate losses $\ell_k=f_T(x_k, \lambda)$ for $k$ between $1$ and $K$
   \STATE Calculate weights $w_1, w_2, \dots, w_K = \operatorname{softmax}([-\ell_1, -\ell_2, \dots, -\ell_K]/\tau)$
   \STATE Calculate $x^* = w_1 x_1 + w_2 x_2 + \dots + w_K x_K$
   \STATE Update $\lambda \leftarrow \lambda - \alpha \frac{\partial f_V(x^*)}{\partial \lambda}$
   \ENDFOR
\end{algorithmic}
\end{algorithm}

\subsection{Rotation transformation}
As part of the rotation transformation problem, we try to prepare a model for the classification of rotated images. We use MNIST images \citep{LeCun1998MNISTDatabase} and train the base model with unrotated training images, while testing is done with images rotated by $30^\circ$. We split the original training set to create a meta-validation set of size 10,000 with images rotated by $30^\circ$.

To prepare the model for the target problem, we meta-learn a rotation transformation alongside training the base model -- which we apply to the unrotated images. Our base model is LeNet \citep{LeCun1989HandwrittenLearning} that has two CNN layers followed by three fully-connected layers. We use ReLU non-linearity and max-pooling. The base model is trained with Adam optimizer \citep{Kingma2015Adam:Optimization} with 0.001 learning rate, while the meta-parameter is optimized with Adam optimizer with learning rate of 0.01. We use a batch size of 128 and cross-entropy loss $\ell$. EvoGrad parameters are $\tau=0.05, \sigma=0.001, K=2$. We sample the noise parameters as $\boldsymbol{\epsilon}_k\sim \sigma \text{sign}(\mathcal{N}(\boldsymbol{0}, \boldsymbol{I}))$, and we use this formulation also in the further practical meta-learning problems -- it is a better-controlled version of simple $\mathcal{N}(\boldsymbol{0}, \sigma\boldsymbol{I})$. We train the models for 5 epochs and repeat the experiments 5 times. The algorithm is summarized in Algorithm \ref{alg:evogradrot}.

Rotations are performed using a model with one learnable parameter $\lambda$ (angle). The input that the model receives is rotated using matrix:
$$\begin{pmatrix}
\cos(\lambda) & -\sin(\lambda)\\
\sin(\lambda) & \phantom{-}\cos(\lambda)
\end{pmatrix}.$$

\begin{algorithm}[h]
   \caption{Training with EvoGrad hypergradient -- rotation transformation}
   \label{alg:evogradrot}
\begin{algorithmic}[1]
   \STATE {\bfseries Input:} $\alpha$: learning rate; $\beta$: meta-learning rate; $\sigma$: noise parameter; $K$: number of model candidates; $\tau$: temperature
   \STATE {\bfseries Output:} $\boldsymbol{\theta}$: trained model; $\lambda$: rotation parameter
   \STATE Initialize $\boldsymbol{\theta} \sim p(\boldsymbol{\theta})$ and $\lambda=0$
   \WHILE{\textit{training}}
   \STATE Sample minibatch of training $x_t, y_t$ (standard) and validation $x_v, y_v$ (rotated) examples
   \STATE Update $\boldsymbol{\theta} \leftarrow \boldsymbol{\theta} - \alpha \nabla_{\boldsymbol{\theta}} \ell(f_{\boldsymbol{\theta}}(f_{\lambda}(x_t)), y_t)$
   \STATE Sample $K$ noise parameters $\boldsymbol{\epsilon}_k\sim \sigma \text{sign}(\mathcal{N}(\boldsymbol{0}, \boldsymbol{I}))$ and use them to create $\boldsymbol{\theta}_k=\boldsymbol{\theta}+\boldsymbol{\epsilon}_k$
   \STATE Calculate losses $\ell_k=\ell(f_{\boldsymbol{\theta}_k}(f_{\lambda}(x_t)), y_t)$ for $k$ between $1$ and $K$
   \STATE Calculate weights $w_1, w_2, \dots, w_K = \operatorname{softmax}([-\ell_1, -\ell_2, \dots, -\ell_K]/\tau)$
   \STATE Calculate $\boldsymbol{\theta}^* = w_1 \boldsymbol{\theta}_1 + w_2 \boldsymbol{\theta}_2 + \dots + w_K \boldsymbol{\theta}_K$
   \STATE Update $\lambda \leftarrow \lambda - \beta \nabla_\lambda\ell(f_{\boldsymbol{\theta}^*}(x_v), y_v)$
   \ENDWHILE
\end{algorithmic}
\end{algorithm}

We compare our meta-learning approach to simple standard training that does not use the rotation transformer. In such case we keep the same settings as before and update the model simply as $\boldsymbol{\theta} \leftarrow \boldsymbol{\theta} - \alpha \nabla_{\boldsymbol{\theta}} \ell(f_{\boldsymbol{\theta}}(x_t), y_t)$. The results prove EvoGrad is capable of meta-learning suitable values.

\subsection{Cross-domain few-shot classification via learned feature-wise transformation}
We extend the \textit{Learning-to-Learn Feature-Wise Transformation} method from \citep{Tseng2020Cross-domainTransformation} to show the practical impact that EvoGrad can make. The goal of the LFT method is to make metric-based few-shot learners robust to domain shift. A detailed description of the LFT method is provided in \citep{Tseng2020Cross-domainTransformation}, and here we describe the main changes that are needed to use EvoGrad for LFT. The key difference is that we do not backpropagate via standard model update that leads to higher memory and time consumption (we measure maximum allocated memory and time per epoch).

We summarize how EvoGrad is applied to LFT in Algorithm \ref{alg:evogradcdfsl}. A metric based model (we choose RelationNet \citep{Sung2018LearningLearning}) includes feature encoder $E_{\boldsymbol{\theta}_e}$ and metric function $M_{\boldsymbol{\theta}_m}$. Feature transformation layers parameterized by ${\boldsymbol{\theta}_f=\{\boldsymbol{\theta}_{\gamma}, \boldsymbol{\theta}_{\beta}\}}$ are integrated into the feature encoder to form $E_{\boldsymbol{\theta}_e, \boldsymbol{\theta}_f}$. Similarly as \citep{Tseng2020Cross-domainTransformation}, we sample pseudo-seen $\mathcal{T}^{\text{ps}}$ and pseudo-unseen $\mathcal{T}^{\text{pu}}$ domains from the seen domains $\left\{\mathcal{T}_{1}^{\text {seen }}, \mathcal{T}_{2}^{\text {seen }}, \cdots, \mathcal{T}_{n}^{\text {seen }}\right\}$. In each step, we sample pseudo-seen and pseudo-unseen few-shot learning tasks that both include support and query examples. The pseudo-seen task is described as $T^{\mathrm{ps}}=\left\{\left(\mathcal{X}_{s}^{\mathrm{ps}}, \mathcal{Y}_{s}^{\mathrm{ps}}\right),\left(\mathcal{X}_{q}^{\mathrm{ps}}, \mathcal{Y}_{q}^{\mathrm{ps}}\right)\right\} \in \mathcal{T}^{\mathrm{ps}}$ and the pseudo-unseen task is $T^{\mathrm{pu}}=\left\{\left(\mathcal{X}_{s}^{\mathrm{pu}}, \mathcal{Y}_{s}^{\mathrm{pu}}\right),\left(\mathcal{X}_{q}^{\mathrm{pu}}, \mathcal{Y}_{q}^{\mathrm{pu}}\right)\right\} \in \mathcal{T}^{\mathrm{pu}}$, for task examples $\mathcal{X}$ with labels $\mathcal{Y}$.

We have used the exact same set-up as \citep{Tseng2020Cross-domainTransformation} with their official implementation (for RelationNet), so we only describe the additional settings that are unique to us. In particular, EvoGrad-specific parameters are $\tau=0.05, K=2, \sigma=0.001$ (we have used $\sigma$ equal to the learning rate). We have used ResNet-10 \citep{He2015DeepRecognition} backbone for direct comparison with \citep{Tseng2020Cross-domainTransformation}. The datasets that we use are processed in the same way as done by \citep{Tseng2020Cross-domainTransformation}, and they are MiniImagenet \citep{Ravi2017OptimizationLearning}, CUB \citep{Welinder2010Caltech-UCSD200}, Cars \citep{Krause20133DCategorization}, Places \citep{Zhou2018Places:Recognition} and Plantae \citep{Horn2018TheDataset}.

In order to use ResNet-34, we have trained a new ResNet-34 baseline model on MiniImagenet \citep{Ravi2017OptimizationLearning} per \citep{Tseng2020Cross-domainTransformation} instructions. We use the same hyperparameters as were used for ResNet-10, which also means that when using fixed feature transformation layers, we use $\theta_\gamma=0.3, \theta_\beta=0.5$. Note that ResNet-34 ran out of memory for the original second-order LFT approach on 5-way 5-shot task with 16 query examples when using standard GPU with 12 GB GPU memory. If we wanted to use this model also for the second-order approach, we would need to decrease the number of examples in the task appropriately. However, with EvoGrad we do not need to make this compromise and overall it means that EvoGrad scales also to problems where the original second-order approach does not scale because of GPU memory limitations.

\begin{algorithm}[h]
   \caption{Learning-to-learn feature-wise transformation -- with EvoGrad}
   \label{alg:evogradcdfsl}
\begin{algorithmic}[1]
   \STATE {\bfseries Input:} $\left\{\mathcal{T}_{1}^{\text {seen }}, \mathcal{T}_{2}^{\text {seen }}, \cdots, \mathcal{T}_{n}^{\text {seen }}\right\}$: seen domains; $\alpha$: learning rate; $\sigma$: noise parameter; $K$: number of model candidates; $\tau$: temperature
   \STATE {\bfseries Output:} $\boldsymbol{\theta}_e$: feature extractor; $\boldsymbol{\theta}_m$: metric learner; $\boldsymbol{\theta}_f$: feature transformation layers
   \STATE Initialize $\boldsymbol{\theta}_e, \boldsymbol{\theta}_m, \boldsymbol{\theta}_f \sim p(\boldsymbol{\theta}_e), p(\boldsymbol{\theta}_m), p(\boldsymbol{\theta}_f)$
   \WHILE{\textit{training}}
   \STATE Randomly sample non-overlapping pseudo-seen $\mathcal{T}^{\text{ps}}$ and pseudo-unseen $\mathcal{T}^{\text{pu}}$ domains from the seen domains
   \STATE Sample a pseudo-seen task $T^{\text{ps}} \in \mathcal{T}^{\text{ps}}$ and a pseudo-unseen task $T^{\text{pu}} \in \mathcal{T}^{\text{pu}}$
   \STATE \textbf{// Standard update of the metric-based model with pseudo-seen task:}
   \STATE Update $\boldsymbol{\theta}_e, \boldsymbol{\theta}_m \leftarrow \boldsymbol{\theta}_e, \boldsymbol{\theta}_m - \alpha \nabla_{\left(\boldsymbol{\theta}_e, \boldsymbol{\theta}_m\right)} \ell\left(M_{\boldsymbol{\theta}_{m}}\left(\mathcal{Y}_{s}^{\mathrm{ps}}, E_{\boldsymbol{\theta}_{e}, \boldsymbol{\theta}_{f}}\left(\mathcal{X}_{s}^{\mathrm{ps}}\right), E_{\boldsymbol{\theta}_{e}, \boldsymbol{\theta}_{f}}\left(\mathcal{X}_{q}^{\mathrm{ps}}\right)\right), \mathcal{Y}_{q}^{\mathrm{ps}}\right)$
   \STATE \textbf{// EvoGrad computations:}
   \STATE Sample $K$ noise parameters $\left\{\boldsymbol{\epsilon}^{(k)}_e, \boldsymbol{\epsilon}^{(k)}_m \right\}_{k=1}^K \sim \sigma \text{sign}(\mathcal{N}(\boldsymbol{0}, \boldsymbol{I}))$
   \STATE Create $\boldsymbol{\theta}^{(k)}_e=\boldsymbol{\theta}_e+\boldsymbol{\epsilon}^{(k)}_e$ and $\boldsymbol{\theta}^{(k)}_m=\boldsymbol{\theta}_m+\boldsymbol{\epsilon}^{(k)}_m$ for $k$ between $1$ and $K$
   \STATE Calculate losses $\ell_k=\ell\left(M_{\boldsymbol{\theta}_{m}^{(k)}}\left(\mathcal{Y}_{s}^{\mathrm{ps}}, E_{\boldsymbol{\theta}_{e}^{(k)}, \boldsymbol{\theta}_{f}}\left(\mathcal{X}_{s}^{\mathrm{ps}}\right), E_{\boldsymbol{\theta}_{e}^{(k)}, \boldsymbol{\theta}_{f}}\left(\mathcal{X}_{q}^{\mathrm{ps}}\right)\right), \mathcal{Y}_{q}^{\mathrm{ps}}\right)$
   \STATE Calculate weights $w_1, w_2, \dots, w_K = \operatorname{softmax}([-\ell_1, -\ell_2, \dots, -\ell_K]/\tau)$
   \STATE Calculate $\boldsymbol{\theta}^*_e = w_1 \boldsymbol{\theta}_e^{(1)} + w_2 \boldsymbol{\theta}_e^{(2)} + \dots + w_K \boldsymbol{\theta}_e^{(K)}$
   \STATE Calculate $\boldsymbol{\theta}^*_m = w_1 \boldsymbol{\theta}_m^{(1)} + w_2 \boldsymbol{\theta}_m^{(2)} + \dots + w_K \boldsymbol{\theta}_m^{(K)}$
   \STATE \textbf{// Update feature-wise transformation layers with pseudo-unseen task:}
   \STATE Update $\boldsymbol{\theta}_f \leftarrow \boldsymbol{\theta}_f - \alpha \nabla_{\boldsymbol{\theta}_f}  \ell\left(M_{\boldsymbol{\theta}^*_m}\left(\mathcal{Y}_{s}^{\mathrm{pu}}, E_{\boldsymbol{\theta}^*_e}\left(\mathcal{X}_{s}^{\mathrm{pu}}\right), E_{\boldsymbol{\theta}^*_e}\left(\mathcal{X}_{q}^{\mathrm{pu}}\right)\right), \mathcal{Y}_{q}^{\mathrm{pu}}\right)$
   \ENDWHILE
\end{algorithmic}
\end{algorithm}

\subsection{Label noise with Meta-Weight-Net}
We use the experimental set-up from \citep{Shu2019Meta-Weight-Net:Weighting} for the label noise experiments, together with their official implementation. The label noise experiments use ResNet-32 model and 60 epochs, each of which has 500 iterations. CIFAR-10 and CIFAR-100 \citep{Krizhevsky2009LearningImages} datasets are used. Meta-Weight-Net is represented by a neural network with two linear layers with hidden size of 300 units, ReLU nonlinearity in between and sigmoid output unit. Meta-Weight-Net weights instance-wise losses for each example in the minibatch, which are then combined together by taking their sum. EvoGrad specific parameters are $\tau=0.05, K=2, \sigma=0.001$. The level of label noise depends on the specific scenario considered -- 40\%, 20\% or 0\%.

We provide an overview of the EvoGrad approach applied to the label noise with Meta-Weight-Net problem in Algorithm \ref{alg:mwn}. Even though we do the standard update using noisy examples after the meta-update, the order could be swapped and we simply follow the order chosen by \citep{Shu2019Meta-Weight-Net:Weighting}. Detailed explanations are provided in \citep{Shu2019Meta-Weight-Net:Weighting}, we only explain how we modify the method to use EvoGrad. Note that we do not rerun the baseline experiments and we directly take the reported values from the Meta-Weight-Net paper \citep{Shu2019Meta-Weight-Net:Weighting}. However, we do our own rerun of standard second-order Meta-Weight-Net to get memory and runtime statistics.

\begin{algorithm}[h]
   \caption{Meta-Weight-Net for label noise -- with EvoGrad}
   \label{alg:mwn}
\begin{algorithmic}[1]
   \STATE {\bfseries Input:} $\alpha$: learning rate; $\sigma$: noise parameter; $K$: number of model candidates; $\tau$: temperature
   \STATE {\bfseries Output:} $\boldsymbol{\theta}$: trained model; $\boldsymbol{\omega}$: Meta-Weight-Net parameters
   \STATE Initialize $\boldsymbol{\theta}, \boldsymbol{\omega} \sim p(\boldsymbol{\theta}), p(\boldsymbol{\omega})$
   \WHILE{\textit{training}}
   \STATE Sample minibatch of training $x_t, y_t$ (noisy) and validation $x_v, y_v$ (clean) examples
   \STATE \textbf{// EvoGrad update:}
   \STATE Sample $K$ noise parameters $\boldsymbol{\epsilon}_k\sim \sigma \text{sign}(\mathcal{N}(\boldsymbol{0}, \boldsymbol{I}))$ and use them to create $\boldsymbol{\theta}_k=\boldsymbol{\theta}+\boldsymbol{\epsilon}_k$
   \STATE Calculate losses $\ell_k=f_{\boldsymbol{\omega}}\left(\ell(f_{\boldsymbol{\theta}_k}(x_t), y_t)\right)$ for $k$ between $1$ and $K$
   \STATE Calculate weights $w_1, w_2, \dots, w_K = \operatorname{softmax}([-\ell_1, -\ell_2, \dots, -\ell_K]/\tau)$
   \STATE Calculate $\boldsymbol{\theta}^* = w_1 \boldsymbol{\theta}_1 + w_2 \boldsymbol{\theta}_2 + \dots + w_K \boldsymbol{\theta}_K$
   \STATE Update $\boldsymbol{\omega} \leftarrow \boldsymbol{\omega} - \alpha \nabla_{\boldsymbol{\omega}}\ell(f_{\boldsymbol{\theta}^*}(x_v), y_v)$
   \STATE \textbf{// Standard update using noisy examples and MWN:}
   \STATE Update $\boldsymbol{\theta} \leftarrow \boldsymbol{\theta} - \alpha \nabla_{\boldsymbol{\theta}} f_{\boldsymbol{\omega}}\left(\ell(f_{\boldsymbol{\theta}}(x_t), y_t)\right)$
   \ENDWHILE
\end{algorithmic}
\end{algorithm}

In addition, we provide further details about Meta-Weight-Net scalability analyses. We have chosen MWN to conduct these analyses because it represents a real problem where meta-learning is helpful, yet the memory consumption and time requirements are small enough to allow us to easily evaluate scaling up of the numbers of parameters. All Meta-Weight-Net scalability experiments are repeated 5 times, but we do not run them fully -- we only do 10 epochs to get estimates of the time per epoch.

\begin{figure}[h!]
  \centering
  \includegraphics[width=\textwidth]{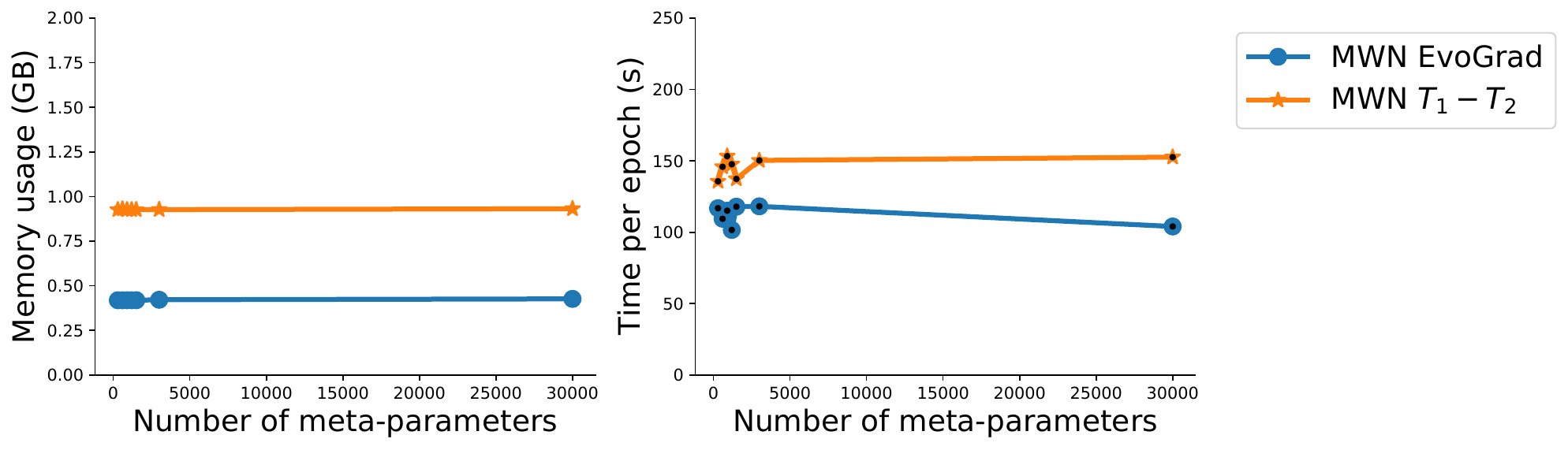}
  \caption{Memory and time scaling of MWN EvoGrad vs original second-order Meta-Weight-Net -- when changing the number of learnable hyperparameters (meta-parameters). The number of meta-parameters does not noticeably influence the memory usage and time per epoch in this case.}
  \label{fig:mwn_plots_meta}
\end{figure}

\begin{figure}[h!]
  \centering
  \includegraphics[width=\textwidth]{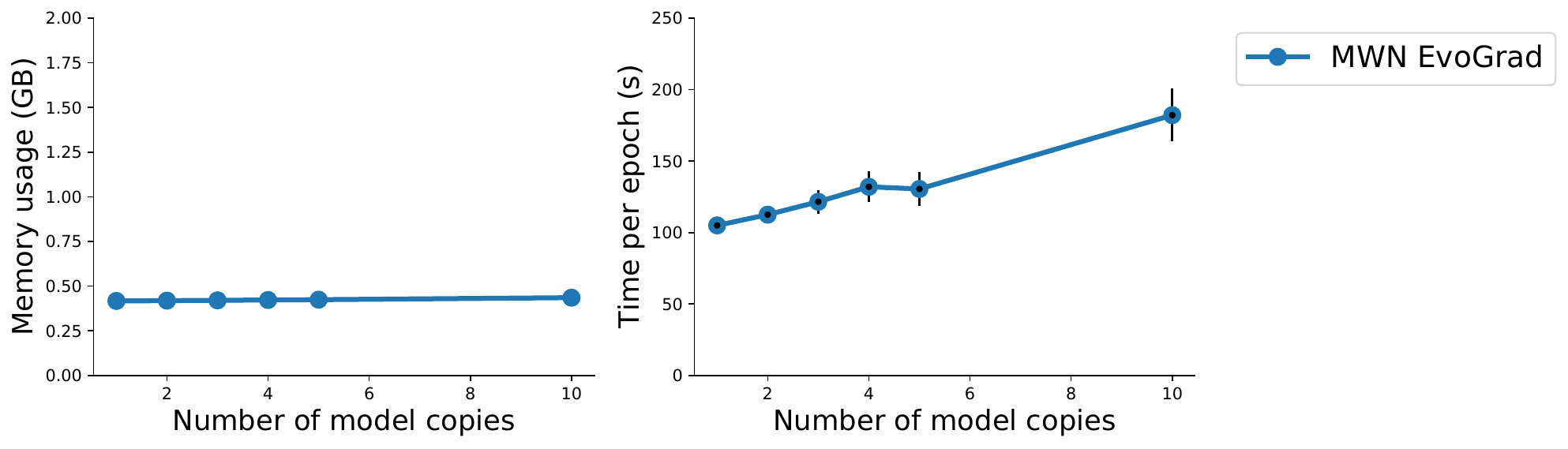}
  \caption{Memory and time scaling of MWN EvoGrad -- when using different numbers of model copies. A larger number of model copies does not increase the memory usage in this case, but it leads to a larger time per epoch.}
  \label{fig:mwn_plots_model_copies}
\end{figure}

We have provided the main results that evaluate the impact of using a model with significantly more parameters in the main part of the paper. Here we provide additional figures. Figure \ref{fig:mwn_plots_meta} shows the impact of variable number of meta-parameters (number of hidden units in MWN). We can see the number of meta-parameters does not significantly impact the memory usage or runtime. This is likely because we use reverse-mode backpropagation that becomes more expensive with more model parameters and not hyperparameters \citep{Micaelli2019Zero-shotMatching}. Further, the number of meta-parameters still remains small compared to the size of the model. Figure \ref{fig:mwn_plots_model_copies} shows the number of model copies does not lead to increased memory consumption, perhaps because we only keep the model weights in memory and not also many intermediate variables like activations that are needed for backpropagation -- backpropagation is significantly more expensive in terms of memory than forward propagation \citep{Rajeswaran2019Meta-learningGradients}. The runtime increases slightly with additional model copies, which comes from the need to calculate additional forward propagations.

\subsection{Low-resource cross-lingual learning with MetaXL}
MetaXL \citep{Xia2021MetaXL:Learning} is an approach that meta-learns meta representation transformation to improve transfer in low-resource cross-lingual learning. We show how EvoGrad is applied to MetaXL in Algorithm \ref{alg:metaxl}

\begin{algorithm}[h]
   \caption{MetaXL for cross-lingual learning -- with EvoGrad}
   \label{alg:metaxl}
\begin{algorithmic}[1]
   \STATE {\bfseries Input:} $\alpha, \beta$: learning rates; $\sigma$: noise parameter; $K$: number of model candidates; $\tau$: temperature; $D_t, D_s$: input data from the target and source language
   \STATE {\bfseries Output:} $\boldsymbol{\theta}$: trained model; $\boldsymbol{\omega}$: representation transformation network
   \STATE Initialize base model parameters $\boldsymbol{\theta}$ with pretrained XLM-R weights, initialize parameters of the
representation transformation network $\boldsymbol{\omega}$ randomly
   \WHILE{\textit{training}}
   \STATE Sample a source batch $(x_s, y_s)$ from $D_s$ and a target batch $(x_t, y_t)$ from $D_t$
   \STATE \textbf{// EvoGrad update:}
   \STATE Sample $K$ noise parameters $\boldsymbol{\epsilon}_k\sim \sigma \text{sign}(\mathcal{N}(\boldsymbol{0}, \boldsymbol{I}))$ and use them to create $\boldsymbol{\theta}_k=\boldsymbol{\theta}+\boldsymbol{\epsilon}_k$
   \STATE Calculate losses $\ell_k=\ell(f_{\boldsymbol{\omega}\circ{\boldsymbol{\theta}_k}}(x_s), y_s)$ for $k$ between $1$ and $K$
   \STATE Calculate weights $w_1, w_2, \dots, w_K = \operatorname{softmax}([-\ell_1, -\ell_2, \dots, -\ell_K]/\tau)$
   \STATE Calculate $\boldsymbol{\theta}^* = w_1 \boldsymbol{\theta}_1 + w_2 \boldsymbol{\theta}_2 + \dots + w_K \boldsymbol{\theta}_K$
   \STATE Update $\boldsymbol{\omega} \leftarrow \boldsymbol{\omega} - \beta \nabla_{\boldsymbol{\omega}}\ell(f_{\boldsymbol{\theta}^*}(x_t), y_t)$
   \STATE \textbf{// Standard update using representation transformation network:}
   \STATE Update $\boldsymbol{\theta} \leftarrow \boldsymbol{\theta} - \alpha \nabla_{\boldsymbol{\theta}} \ell(f_{\boldsymbol{\omega}\circ{\boldsymbol{\theta}}}(x_s), y_s)$
   \ENDWHILE
\end{algorithmic}
\end{algorithm}

In order to do experiments, we have taken the official code provided by \citep{Xia2021MetaXL:Learning} and tried to replicate their experiments as closely as possible. We used the named entity recognition (NER) task with English source language. The only change we made is a smaller batch size: 12 instead of 16 to fit into the memory of the largest GPUs that we have currently available. All details are described in \citep{Xia2021MetaXL:Learning}. For EvoGrad we have selected the same hyperparameters as for the other tasks in our paper (two model candidates, $\sigma=0.001$ and $\tau=0.05$). In order to make the implementation of EvoGrad on MetaXL simple, we have only applied noise perturbation on the top layer of the model. It is likely that in practice it is enough to only apply the noise to the top layer, which can make using EvoGrad very simple in most cases.

\subsection{Datasets availability}
All datasets that we use are freely available and their details are described in \citep{Tseng2020Cross-domainTransformation}, \citep{Shu2019Meta-Weight-Net:Weighting} and \citep{Xia2021MetaXL:Learning} -- including how to download them.

\subsection{Computational resources}
Illustration using a 1-dimensional problem and rotation transformation can be easily run on a laptop GPU. For cross-domain few-shot learning with LFT and label noise with MWN, we have used an internal cluster with NVIDIA GPUs - Titan X or P100 (all with 12GB GPU memory). For MetaXL we have used NVIDIA 3090 Ti GPUs with 24GB memory. When reporting the time and memory statistics we made sure to use the same model of GPU so that the comparisons are accurate. The experiments were allocated 14 GB RAM memory and 6 CPUs to allow for faster data loading (fewer resources would also be suitable).

\section{Evaluation of hypernetworks}
We have evaluated hypernetworks \citep{Lorraine2018StochasticHypernetworks} for cross-domain few-shot classification via learned feature-wise transformation, to find if the approach can be useful for recent meta-learning applications. To make the approach computationally viable, we have used hypernetworks with a bottleneck. For $H$ hyperparameters, $P$ model parameters and bottleneck size of $B$, our hypernetwork $\boldsymbol{\phi}$ consists of two layers, one with a weight matrix of $H\times B$, followed by $B \times P$ weight matrix, with sigmoid non-linearity in between. Note that $B$ needs to be relatively small and directly using one layer with a weight matrix of $H \times P$ would require far more memory than normally available -- for the considered problem. Following \citep{Lorraine2018StochasticHypernetworks}, we have used bottleneck size $B=10$. We have used the exact same experimental set-up as in our other experiments.

Our results in Table \ref{tab:lfthyper} show hypernetworks fail to discover a good solution within the standard number of iterations used throughout, and their performance is poor. The results highlight that generating model parameters based on the hyperparameters may not be sufficient in more challenging and more realistic meta-learning problems. It also explains why hypernetworks are not commonly used in meta-learning applications.

\begin{table}[h!]
\caption{RelationNet test accuracies (\%) and 95\% confidence intervals across test tasks on various unseen datasets. 5-way 1-shot learning at the top and 5-way 5-shot learning at the bottom. Hypernetworks lead to significantly worse accuracies than $T_1-T_2$ and EvoGrad, showing they fail to generate well-performing model parameters.}
\label{tab:lfthyper}
\centering
\begin{tabular}{lcccc}
\toprule
Scenario & CUB & Cars & Places & Plantae \\
\midrule
LFT with hypernetworks & 38.94 $\pm$ 0.57 & 30.10 $\pm$ 0.48 & 38.07 $\pm$ 0.58 & 33.83 $\pm$ 0.58 \\
LFT with $T_1-T_2$ & 46.03 $\pm$ 0.60 & 31.50 $\pm$ 0.49 & 49.29 $\pm$ 0.65 & 36.34 $\pm$ 0.59 \\
LFT with EvoGrad & 47.39 $\pm$ 0.61 & 32.51 $\pm$ 0.56 & 50.70 $\pm$ 0.66 & 36.00 $\pm$ 0.56 \\
\midrule
LFT with hypernetworks & 56.91 $\pm$ 0.57 & 40.64 $\pm$ 0.56 & 56.08 $\pm$ 0.58 & 44.73 $\pm$ 0.57 \\
LFT with $T_1-T_2$ & 65.94 $\pm$ 0.56 & 43.88 $\pm$ 0.56 & 65.57 $\pm$ 0.57 & 51.43 $\pm$ 0.55 \\
LFT with EvoGrad & 64.63 $\pm$ 0.56 & 42.64 $\pm$ 0.58 & 66.54 $\pm$ 0.57 & 52.92 $\pm$ 0.57 \\
\bottomrule
\end{tabular}
\end{table}

\section{Comparison to more meta-learning approaches}
In this section we provide an extended comparison of hypergradient approximations by various gradient-based meta-learners, similar to the analysis done in \citep{Lorraine2020OptimizingDifferentiation}. The approximations themselves are provided in Table \ref{tab:hypergradcomparisonfull}, while the time and memory requirements are given in Table \ref{tab:memorycomparisonfull}.

\begin{table}[h!]
\caption{Comparison of hypergradient approximations of different gradient-based meta-learning methods. Number of inner-loop steps is denoted by $i$. Note that also one-step approximation methods can be used once per $i$ steps. $\boldsymbol{\theta}^*$ describes the optimal model parameters given $\boldsymbol{\lambda}$, while $\widehat{\boldsymbol{\theta}^*}$ represents their approximation.}
\label{tab:hypergradcomparisonfull}
\centering
\begin{tabular}{ll}
\toprule
Method & Hypergradient approximation \\
\midrule
Unrolled diff. \citep{Maclaurin2015Gradient-basedLearning} & $\frac{\partial \ell_V}{\partial \boldsymbol{\lambda}}-\frac{\partial \ell_V}{\partial \boldsymbol{\theta}} \times\left. \sum_{j \leq i}\left[\prod_{k<j} I-\left.\frac{\partial^{2} \ell_T}{\partial \boldsymbol{\theta} \partial \boldsymbol{\theta}^{T}}\right|_{\boldsymbol{\theta}_{i-k}}\right] \frac{\partial^{2} \ell_T}{\partial \boldsymbol{\theta} \partial \boldsymbol{\lambda}^{T}}\right|_{\boldsymbol{\theta}_{i-j}}$\\
\parbox[t]{3cm}{$K$-step truncated\\unrolled diff. \citep{Shaban2019TruncatedOptimization}}
& $\frac{\partial \ell_V}{\partial \boldsymbol{\lambda}}-\frac{\partial \ell_V}{\partial \boldsymbol{\theta}} \times\left.\sum_{K \leq j \leq i}\left[\prod_{k<j} I-\left.\frac{\partial^{2} \ell_T}{\partial \boldsymbol{\theta} \partial \boldsymbol{\theta}^{T}}\right|_{\boldsymbol{\theta}_{i-k}}\right] \frac{\partial^{2} \ell_T}{\partial \boldsymbol{\theta} \partial \boldsymbol{\lambda}^{T}}\right|_{\boldsymbol{\theta}_{i-j}}$ \\ \\
$T_1-T_2$ \citep{Luketina2016ScalableHyperparameters}&  $\frac{\partial \ell_V}{\partial \boldsymbol{\lambda}}-\frac{\partial \ell_V}{\partial \boldsymbol{\theta}} \times\left.[I]^{-1} \frac{\partial^{2} \ell_T}{\partial \boldsymbol{\theta} \partial \boldsymbol{\lambda}^{T}}\right|_{\widehat{\boldsymbol{\theta}^{*}}(\boldsymbol{\lambda})}$\\ 
Hypernetworks \citep{Lorraine2018StochasticHypernetworks}&  $\frac{\partial \ell_V}{\partial \boldsymbol{\lambda}}+\frac{\partial \ell_V}{\partial \boldsymbol{\theta}} \times \frac{\partial \boldsymbol{\theta}^*_{\boldsymbol{\phi}}}{\partial \boldsymbol{\lambda}}$ where $\boldsymbol{\theta}^*_{\boldsymbol{\phi}} (\boldsymbol{\lambda}) = \argmin_{\boldsymbol{\phi}} \ell_T\left(\boldsymbol{\lambda}, \boldsymbol{\theta}_{\boldsymbol{\phi}} (\boldsymbol{\lambda})\right)$ \\
Exact IFT \citep{Lorraine2020OptimizingDifferentiation} & $\frac{\partial \ell_V}{\partial \boldsymbol{\lambda}}-\frac{\partial \ell_V}{\partial \boldsymbol{\theta}} \times\left.\left[\frac{\partial^{2} \ell_T}{\partial \boldsymbol{\theta} \partial \boldsymbol{\theta}^{T}}\right]^{-1} \frac{\partial^{2} \ell_T}{\partial \boldsymbol{\theta} \partial \boldsymbol{\lambda}^{T}}\right|_{\boldsymbol{\theta}^{*}(\boldsymbol{\lambda})}$\\
Neumann IFT \citep{Lorraine2020OptimizingDifferentiation} & $\frac{\partial \ell_V}{\partial \boldsymbol{\lambda}}-\frac{\partial \ell_V}{\partial \boldsymbol{\theta}} \times\left.\left(\sum_{j<i}\left[I-\frac{\partial^{2} \ell_T}{\partial \boldsymbol{\theta} \partial \boldsymbol{\theta}^{T}}\right]^{j}\right) \frac{\partial^{2} \ell_T}{\partial \boldsymbol{\theta} \partial \boldsymbol{\lambda}^{T}}\right|_{\widehat{\boldsymbol{\theta}^{*}}(\boldsymbol{\lambda})}$\\
EvoGrad (ours) & $\frac{\partial \ell_V}{\partial \boldsymbol{\lambda}}+\frac{\partial \ell_{V}}{\partial \boldsymbol{\theta}} \times\left. \mathcal{E}  \frac{\partial \boldsymbol{w}}{\partial \boldsymbol{\ell}} \frac{\partial  \boldsymbol{\ell}}{\partial \boldsymbol{\lambda}} = \frac{\partial \ell_V}{\partial \boldsymbol{\lambda}}+\frac{\partial \ell_V}{\partial \boldsymbol{\theta}} \times \mathcal{E} \frac{\partial \operatorname{softmax}(-\boldsymbol{\ell})}{\boldsymbol{\partial\lambda}}\right|_{\widehat{\boldsymbol{\theta}^{*}}(\boldsymbol{\lambda})}$ \\
\bottomrule
\end{tabular}
\end{table}

\begin{table}[h!]
\caption{Comparison of asymptotic time and memory requirements of EvoGrad and other gradient-based meta-learners. $P$ is the number of model parameters, $H$ is the number of hyperparameters, $I$ is the number of inner-loop steps, $N$ is the number of model copies in EvoGrad. Note this is a first-principles analysis, so the time requirements are different when using e.g. reverse-mode backpropagation that uses parallelization.}
\label{tab:memorycomparisonfull}
\centering
\begin{tabular}{lcc}
\toprule
Method & Time requirements & Memory requirements \\
\midrule
Unrolled diff. \citep{Maclaurin2015Gradient-basedLearning} & $\mathcal{O}(IP^2+PH)$ & $\mathcal{O}(PI+H)$\\
$K$-step truncated unrolled diff. \citep{Shaban2019TruncatedOptimization} & $\mathcal{O}(KP^2+PH)$ & $\mathcal{O}(PK+H)$\\
$T_1-T_2$ \citep{Luketina2016ScalableHyperparameters}& $\mathcal{O}(PH)$ & $\mathcal{O}(P+H)$ \\ 
Linear hypernetworks \citep{Lorraine2018StochasticHypernetworks}& $\mathcal{O}(PH)$ & $\mathcal{O}(PH)$ \\ 
Neumann IFT \citep{Lorraine2020OptimizingDifferentiation} & $\mathcal{O}(P^2+PH)$  & $\mathcal{O}(P+H)$\\
EvoGrad (ours) & $\mathcal{O}(NP+H)$ & $\mathcal{O}(P+H)$\\
\bottomrule
\end{tabular}
\end{table}

\end{document}